\documentclass[10pt,journal,compsoc]{IEEEtran}
\usepackage{xspace}

\usepackage{times}
\usepackage{epsfig}
\usepackage{graphicx}
\usepackage{amsmath}
\usepackage{amssymb}
\usepackage{cite}
\usepackage{multirow}
\usepackage{booktabs,tabularx}
\usepackage{enumitem}
\usepackage{makecell}
\usepackage{indentfirst}
\usepackage{siunitx}
\usepackage{algorithm}
\usepackage{algorithmic}
\usepackage{subfigure}
\usepackage{hyperref}
\usepackage[switch]{lineno}

\makeatletter
\DeclareRobustCommand\onedot{\futurelet\@let@token\@onedot}
\def\@onedot{\ifx\@let@token.\else.\null\fi\xspace}

\def\ie{\emph{i.e}\onedot}

\def\etal{\emph{et al}\onedot}
\makeatother

\ifCLASSINFOpdf
\else
\fi
\begin{document}
\title{Partial Scene Text Retrieval}

\author{
Hao Wang\textsuperscript{\rm 1}\thanks{ {$\bullet$ This work was done during H. Wang's internship at Huawei.}},
Minghui~Liao\textsuperscript{\rm 2},
Zhouyi~Xie\textsuperscript{\rm 2},
Wenyu Liu\textsuperscript{\rm 1},
Xiang Bai\textsuperscript{\rm 1}\thanks{{$\bullet$  X. Bai is the corresponding author and is with the School of Software Engineering, Huazhong University of Science and Technology, Wuhan, 430074, China.}}\\
\textsuperscript{\rm 1}Huazhong University of Science and Technology,
\textsuperscript{\rm 2}Huawei Inc.\\
{\tt\small \{xbai,liuwy\}@hust.edu.cn, \{wanghao687,liaominghui1\}@huawei.com}\\
}



\IEEEtitleabstractindextext{%
\begin{abstract}
The task of partial scene text retrieval involves localizing and searching for text instances that are the same or similar to a given query text from an image gallery. However, existing methods can only handle text-line instances, leaving the problem of searching for partial patches within these text-line instances unsolved due to a lack of patch annotations in the training data. To address this issue, we propose a network that can simultaneously retrieve both text-line instances and their partial patches. Our method embeds the two types of data (query text and scene text instances) into a shared feature space and measures their cross-modal similarities. To handle partial patches, our proposed approach adopts a Multiple Instance Learning (MIL) approach to learn their similarities with query text, without requiring extra annotations. However, constructing bags, which is a standard step of conventional MIL approaches, can introduce numerous noisy samples for training, and lower inference speed. To address this issue, we propose a Ranking MIL (RankMIL) approach to adaptively filter those noisy samples. Additionally, we present a Dynamic Partial Match Algorithm (DPMA) that can directly search for the target partial patch from a text-line instance during the inference stage, without requiring bags. This greatly improves the search efficiency and the performance of retrieving partial patches.
We evaluate the proposed method on both English and Chinese datasets in two tasks: retrieving text-line instances and partial patches. For English text retrieval, our method outperforms state-of-the-art approaches by 8.04\% mAP and 12.71\% mAP on average, respectively, among three datasets for the two tasks. For Chinese text retrieval, our approach surpasses state-of-the-art approaches by 24.45\% mAP and 38.06\% mAP on average, respectively, among three datasets for the two tasks. {The source code and dataset are available at https://github.com/lanfeng4659/PSTR.} 
\end{abstract}

}

\maketitle

\IEEEdisplaynontitleabstractindextext

%
\IEEEpeerreviewmaketitle

\IEEEraisesectionheading{\section{Introduction}\label{sec:introduction}}

In recent years, scene text understanding has attracted significant research interest from the computer vision community due to a large amount of everyday scene images that contain texts. 
Scene text retrieval, introduced by Mishra~\etal~\cite{MishraAJ13}, aims to search for all scene text instances that are the same or similar to given query text from a collection of natural images. Apart from handling scene text instances, partial scene text retrieval further probes their partial patches. Such a task is quite valuable in many applications, such as book search in libraries~\cite{YangHHOZKG17}, key frame extraction of videos~\cite{SongWHXHJ19}, and visual search~\cite{BaiYLXL18,Wang_2022_CVPR}.

Recent text retrieval methods~\cite{GomezECCV20118YOLO+PHOC,mafla2020real, wen2023visual} built on deep learning frameworks extract Pyramidal Histogram Of Character (PHOC)~\cite{AlmazanGFV14} or global features of scene text for measuring distances to query text. However, these methods lack local features of scene text, which prevents them from searching for partial patches. Another feasible solution to scene text retrieval is based on an end-to-end text recognition system, such as~\cite{jaderberg2016reading,HeCVPR2018}. Under this setting, the retrieval results are determined according to the occurrences of the query text within the spotted words. Such methods often achieve unsatisfactory retrieval performance and fall into a local optimum, as they are optimized with a different evaluation metric that requires high accuracy on both detection and recognition.
Thus, when the model produces missed detections or incorrectly spotted words, the target text instance becomes unsearchable.

\begin{figure}[t]
    \includegraphics[width=0.98\linewidth]{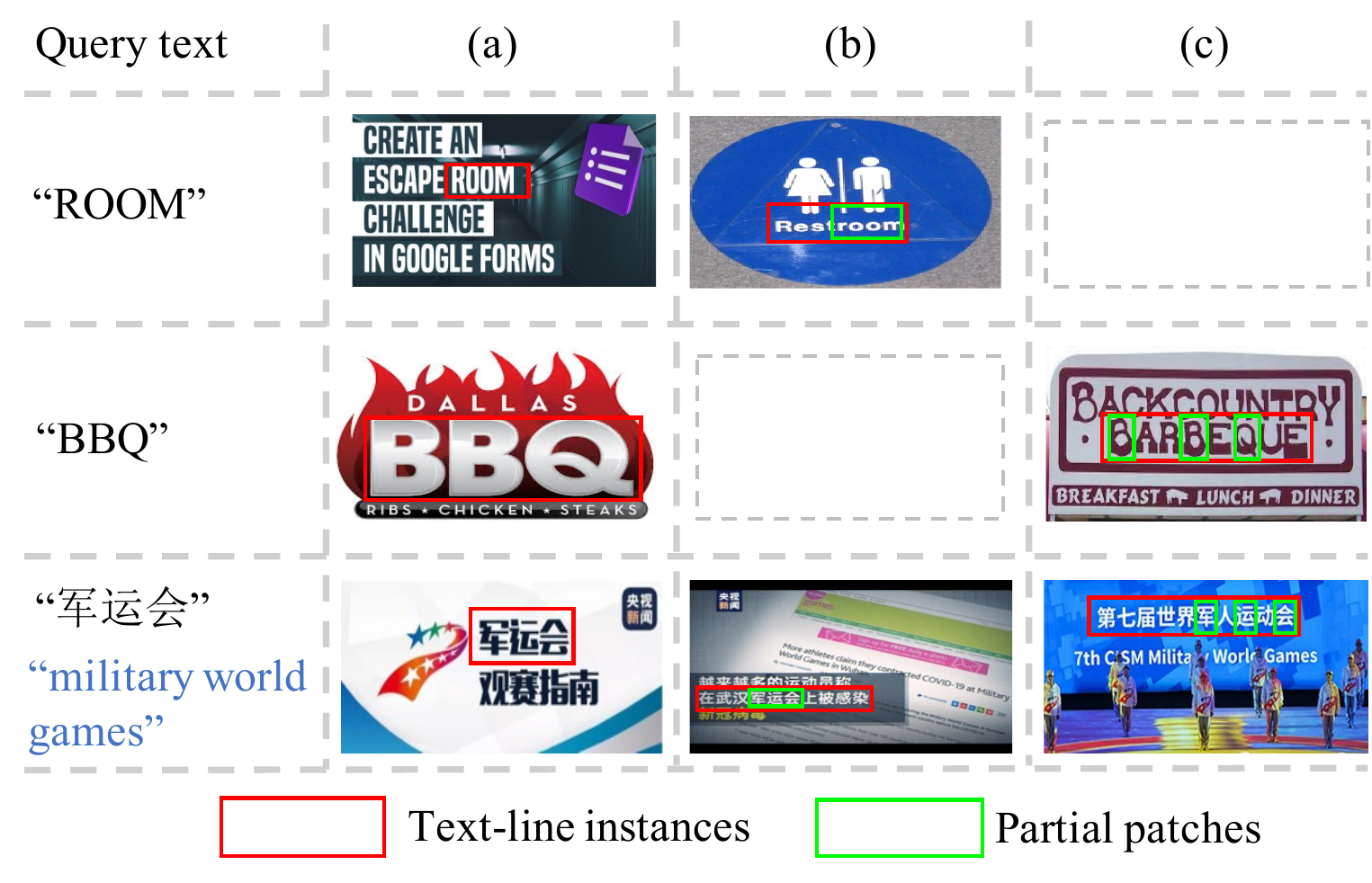}
    \vspace{-2ex}
\centering
  \caption{Result examples of retrieving scene text instances. The target text instances cover various types of instances, such as text-line instances (a), continuous partial patches (b), and non-continuous partial patches (c). The word in blue is the English translation corresponding to the Chinese word in black.}
\vspace{-3ex}  
  \label{abstract}
\end{figure}

This paper proposes a novel method for partial scene text retrieval that is capable of retrieving both text-line instances, as shown in Fig.~\ref{abstract} (a), and their partial patches, as shown in Fig.~\ref{abstract} (b) and (c). We refer to the tasks of retrieving text-line instances and partial patches as Text-line Instance Retrieval (TIR) and Partial Patches Retrieval (PPR), respectively.
Our approach is based on the idea of embedding query text and scene text instances into a shared feature space and measuring their cross-modal similarities in this space. To achieve this, we train the model to learn the similarities between query text and scene text instances using their normalized edit distances, which are commonly used in string matching. However, while training data includes precise annotations of text-line instances, it lacks annotations for partial patches, which makes it difficult to detect them and measure their similarity. Therefore, the supervised learning strategy used for the TIR task cannot be directly applied to the PPR task.

Despite the lack of translation labels for partial patches, it is feasible to determine whether a patch translated as a specific query text is located within a text-line instance by checking whether the translation of the text-line instance contains the query text. This enables us to measure similarities between query text and partial patches using a Multiple Instance Learning (MIL) method. Specifically, we follow the approach proposed in~\cite{conf/nips/ViolaPZ05} and sample partial patches from a text-line instance to form a bag, where the bag label indicates whether the query text exists within the bag. Similarities between partial patches and query text are learned from the constructed bag data.
Unlike general objects such as persons and cars, scene text is a kind of sequence-like object. An image patch of a general object has its discriminative features, but the ones of a scene text usually associate with another word. 
This makes it challenging to precisely determine whether the bag contains the target instance based on its label, resulting in noisy data that would significantly hurt the accuracy of the model.  
However, for the query text included in a text-line instance, we observe that the bag always contains a patch that is more similar to the query text than this text-line instance. Therefore, we propose a Ranking MIL (RankMIL) approach, which ensures that the similarity between the query text and the patch in the bag is greater than the similarity between the query text and the text-line instance.

In the inference stage, for the PPR task, MIL method constructs a bag from each text-line instance and searches for the target instance by calculating the pairwise similarities between features of query text and all patches within the bag.
However, each bag needs many patch\footnote{more than 50 patches on average for each text-line instance on the ReCTS dataset.} to guarantee that at least one patch can cover the target instance, making the search process time-consuming. 
To overcome this challenge, we propose an approach called the Dynamic Partial Match Algorithm (DPMA), which eliminates the need to construct a bag. Instead, it directly obtains the feature of the target instance from the feature of the text-line instance. Specifically, the DPMA represents query text and text-line instances as sequential features, consisting of multiple local features. 
It then selects a set of local features of the text-line instance to form a new feature by a dynamic programming algorithm. The selection strategy of this algorithm is to maximize the similarity between the feature of the query text and the selected feature. Since the selected local features are not necessarily spatially adjacent, the DPMA can handle both continuous partial patches (as in Fig.~\ref{abstract} (b)) and non-continuous partial patches (as in Fig.~\ref{abstract} (c)). As a result, the DPMA not only improves inference efficiency but also enhances the performance of the PPR task.

This paper is an extension of its conference version~\cite{Wang_2021_CVPR}. It broadens the application scope of the conference version to the PPR task, achieved from two aspects. Firstly, the RankMIL training strategy is presented for better optimizing the model to reduce accuracy loss brought by the absence of partial patch annotations. Secondly, we propose the DPMA that dynamically searches for partial patches from a text-line instance. This algorithm significantly improves search efficiency and PPR performance.

The contributions in this work are four-fold.
\begin{enumerate}
\item 
We first introduce a novel end-to-end trainable network for learning cross-modal similarities between query text and scene text instances. This network allows for efficient searching of text instances containing a query text from natural images in both Latin and non-Latin scripts.
\item
We, for the first time, discuss the partial scene text retrieval and optimize it by a MIL method. Moreover, we present the RankMIL algorithm for better optimization, significantly improving the performance of the PPR task. Experimental results verify its superiority to the conventional MIL.
\item
The proposed DPMA enhances search efficiency and can handle not only continuous partial patches but also non-continuous ones without requiring additional data or model parameters. 
\item
We collect and annotate a new dataset {{CSVTRv2}} for Chinese scene text retrieval, consisting of 53 pre-defined query words and 3400 Chinese scene text images. This dataset is adopted to verify the effectiveness of text retrieval methods over non-Latin scripts.
\end{enumerate}

\begin{figure*}[t]
    \includegraphics[width=0.95\linewidth]{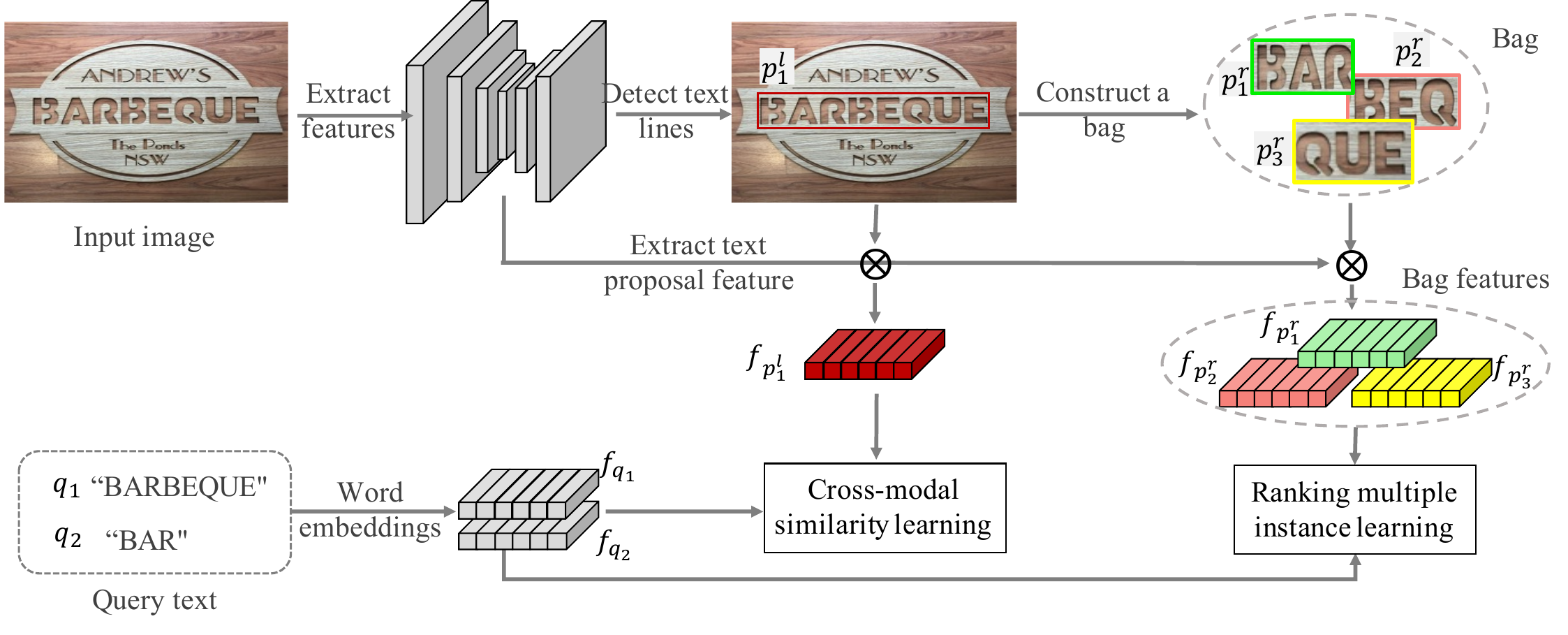}
\centering
\vspace{-2ex}
  \caption{The training phase of our proposed framework. Given an image, text-line proposals are detected, and a bag is constructed within text-line instances.The features of text-line proposals are extracted to the cross-modal similarity learning for the TIR task. Meanwhile, features of instances within the bag are extracted to ranking multiple instance learning for the PPR task. For visualization simplification, we only show the training process of one text-line instance.}
\vspace{-2ex}  
  \label{framework}
\end{figure*}

\section{Related work}
\subsection{Scene Text Retrieval}
Traditional text retrieval methods~\cite{AldavertRTL13, AlmazanGFV14} are proposed to retrieve cropped document text images. These methods typically involve representing both the query text and word images and then calculating the distances between their representations. In method~\cite{AlmazanGFV14}, the Pyramid of Histograms of Characters (PHOC) is first proposed to represent text string for retrieval. Sudholt~\etal~\cite{sudholt2016phocnet} and Wilkinson~\etal~\cite{WilkinsonB16} respectively propose to predict PHOC and DCToW from word image using neural networks. 
In contrast to these methods, Gomez~\etal~\cite{Gomez-BigordaRK17} propose a method that does not rely on hand-crafted representations of text. Instead, they learn the Levenshtein edit distance~\cite{levenshtein1966binary} between the text string and word image directly, which is used to rank word images.
However, the above methods consider perfectly cropped word images as inputs, rather than more universal and challenging scene text images.

Mishra~\etal~\cite{MishraAJ13} first introduce the task of scene text retrieval and adopt two separate steps for character detection and classification. The images are then ranked by scores, which represent the probability of query text existing in the images.  Ghosh~\etal~\cite{GhoshBKV15} propose a scene text retrieval method that uses a selective search algorithm to generate text regions and an SVM-based classifier to predict corresponding PHOCs.
Nevertheless, these methods ignore the relevance and complementarity between text detection and text retrieval, which can lead to suboptimal performance. Additionally, it is inefficient to separate scene text retrieval into two sub-tasks. To integrate text detection and text retrieval in a unified network, Gomez~\etal~\cite{GomezECCV20118YOLO+PHOC} introduce the first end-to-end trainable network for scene text retrieval, where all text instances are represented with PHOCs. Specifically, the method simultaneously predicts text proposals and corresponding PHOCs.
Then, images in the gallery are ranked according to the distance between the PHOC of a query word and the predicted PHOC of each detected text proposal. Some text spotting methods~\cite{HeCVPR2018,jaderberg2016reading} are also adopted for scene text retrieval task. These methods first detect and recognize all possible words in each scene image. Then, the probability that the image contains the given query word is represented as the occurrences of the query word within those spotted words. The above methods suffer from handling partial patches.

\subsection{Multiple Instance Learning}
Multiple Instance Learning (MIL), first proposed for drug activity predition~\cite{journals/ai/DietterichLL97}, is a weakly supervised learning problem. In MIL, a set of bags and their associated bag labels are provided, where each bag contains a collection of instances. The bag label only indicates whether the bag includes a certain kind of instance but does not specify which one. MIL has many applications to computer vision, such as image classification~\cite{conf/cvpr/LiWT13,journals/tip/TangWFL17}, weakly supervised segmentation~\cite{journals/corr/PathakSLD14}, object detection~\cite{conf/nips/ViolaPZ05}. The approach of treating partial patches within the same text-line instance as bags is inspired by ~\cite{conf/nips/ViolaPZ05} and ~\cite{conf/cvpr/OquabBLS15}. The method~\cite{conf/nips/ViolaPZ05} trains MIL for patches around groundtruth locations. Oquab~\etal~\cite{conf/cvpr/OquabBLS15} train a CNN network using the max pooling MIL strategy to localize objects. These MIL-based methods are designed for general objects. However, these MIL-based methods are designed for general objects, while scene text is a sequence-like object, and a patch of text instance often belongs to another word, resulting in noisy training samples. Our presented RankMIL integrates a metric learning to the conventional MIL, which can adaptively filter the noisy samples.

\subsection{Shape Representations for Scene Text}

Shape representations for scene text develop rapidly in scene text detection, from horizontal rectangles to arbitrary polygons. The early scene text detection methods, such as TextBoxes~\cite{conf/aaai/LiaoSBWL17} and CTPN~\cite{conf/eccv/TianHHH016}, mainly focus on horizontal text detection, where horizontal rectangles are adopted to describe the location of the text instances. Then, multi-oriented text detection methods propose more accurate representations for scene text, including rotated rectangles~\cite{conf/cvpr/ZhouYWWZHL17, conf/cvpr/HeLYZTCYWB21} and quadrilaterals~\cite{journals/tip/LiaoSB18, conf/cvpr/LiaoZSXB18}.
Polygon representations are mainly applied in the segmentation methods. PSENet~\cite{conf/cvpr/WangXLHLY019} proposes a progressive scale expansion module for accurate text instance segmentation. DBNet series~\cite{liao2020real, journals/corr/abs-2202-10304} improve the text instance segmentation and simplify the post-processing algorithms of the segmentation-based scene text detectors by introducing a differentiable binarization module.

Recently, some text spotters have introduced novel representations to formulate the shapes of irregular texts. Segmentation-based methods~\cite{qin2019towards,maskv2} detect arbitrarily shaped texts by segmenting their shape masks. 
In methods~\cite{TextPerceptron,conf/eccv/BaekSBPLNL20}, they predict various text components such as character segments and linkages, requiring complex post-processing to restore boundary points of text for feature extraction. ABCNet~\cite{liu2020abcnet} fits arbitrary-shaped text by a parameterized Bezier curve, and the boundary points are sampled from this curve. Wang~\etal~\cite{boundary} propose a simple yet effective pipeline that regresses the boundary points directly. Considering the frequent operator that cuts the detected text-line instance into segments, we adopt boundary points~\cite{boundary} to formulate arbitrary-shaped texts.

\section{Methodology}
The training pipeline is illustrated in Fig.~\ref{framework}. The input of network includes an image and query words $Q = \{q_{i}\}_{i=1}^{|Q|}$, where the operator $|\Psi|$ denotes the number of elements in the set $\Psi$. {Note that, the symbol $``\Psi"$ can represent any set.}
The words in $Q$ are embedded as feature $\{f_{q_i}\}_{i=1}^{|Q|}$ by a word embedding module.
For each image, we obtain text-line proposals $P^l = \{p^l_i\}_{i=1}^{|P^l|}$. Then, we construct a bag $P^r = \{p^r_i\}_{i=1}^{|P^r|}$ from each text-line proposal. The features in $P^l$ are extracted to the cross-modal similarity learning for optimizing the TIR task. Similarly, the features in $P^r$ are extracted to the RankMIL for optimizing the PPR task. 

In this section, first, we introduce the architecture of our network. Second, we give the details of the cross-modal similarity learning for the TIR task. Third, we present the idea of applying a conventional MIL to the PPR task. Fourth, we propose RankMIL for improving the effectiveness of MIL. Then, we present the DPMA to improve the retrieving speed and deal with non-continuous partial instances. Last, we provide the optimization of the whole network.

\subsection{Architecture}
The network consists of a Feature Pyramid Network~\cite{lin2017feature} for extracting feature maps, a text-line proposal module for generating proposals of text-line instances, and a word embedding module for embedding the query words to features.

\subsubsection{Word Embedding Module}\label{sec:wem}
The input query words are a set of text strings that are unable to be directly forwarded by the neural network. So a word embedding module is employed to convey query words into features. There are many word embedding methods, such as Glove~\cite{conf/emnlp/PenningtonSM14} and Word2Vec~\cite{journals/corr/abs-1301-3781}, in the natural language process. Those word embedding methods concentrate on modeling the contextual information of a word among a sentence while distinguishing similar words requires capturing the difference between words.

The word embedding module consists of an embedding layer, a bilinear interpolation operator, and a bidirectional LSTM~\cite{hochreiter1997long} layer. For each word $q_i$ in $Q$, this module outputs a sequence features $f_{q_i} \in \mathbb{R}^{T \times C}$, where the parameters $T$ and $C$ denotes the sequence length and feature dimension. Specifically, $q_{i}$ can be considered as a character sequence. The embedding layer first converts each character into $C$ dimensional feature, producing an embedded feature sequence with size $|q_i| \times C$ ($|q_i|$ is the character number of $q_i$). Then, the feature sequence is concatenated and interpolated as a fixed-length feature with size $T \times C$. Finally, a bidirectional LSTM is used to enhance the contextual information of word features. In our method, the parameters $T$ and $C$ are empirically set to 15 and 128, respectively.

\subsubsection{Text-line Proposal Module}~\label{sec:bpm}

The text-line proposal module aims at generating text-line proposals of arbitrary-shaped texts for the TIR task. 
This module is based on an anchor-free detector, FCOS~\cite{tian2019fcos}, and adapts its prediction heads for text detection. 
For formulating proposals of arbitrary-shaped scene texts, we replace the horizontal box representation with boundary point representation~\cite{boundary}. 
Therefore, there are three heads,~\ie, classification head, centerness head, and boundary head. For each feature map from the output of FPN~\cite{lin2017feature}, the detection is densely predicted on the output feature maps of 4 stacked convolution layers with a stride of 1, padding of 1, and 3×3 kernels. Combing all prediction maps from heads, this module outputs the detected text proposals. 

\textbf{Classification/Centerness Heads} The two heads are the same as the ones in the method~\cite{liu2020abcnet}. The channel of the output feature map in the classification head is 2, as this head only identifies whether each pixel is inside the text instance.

\begin{figure}[t]
    \includegraphics[width=0.9\linewidth]{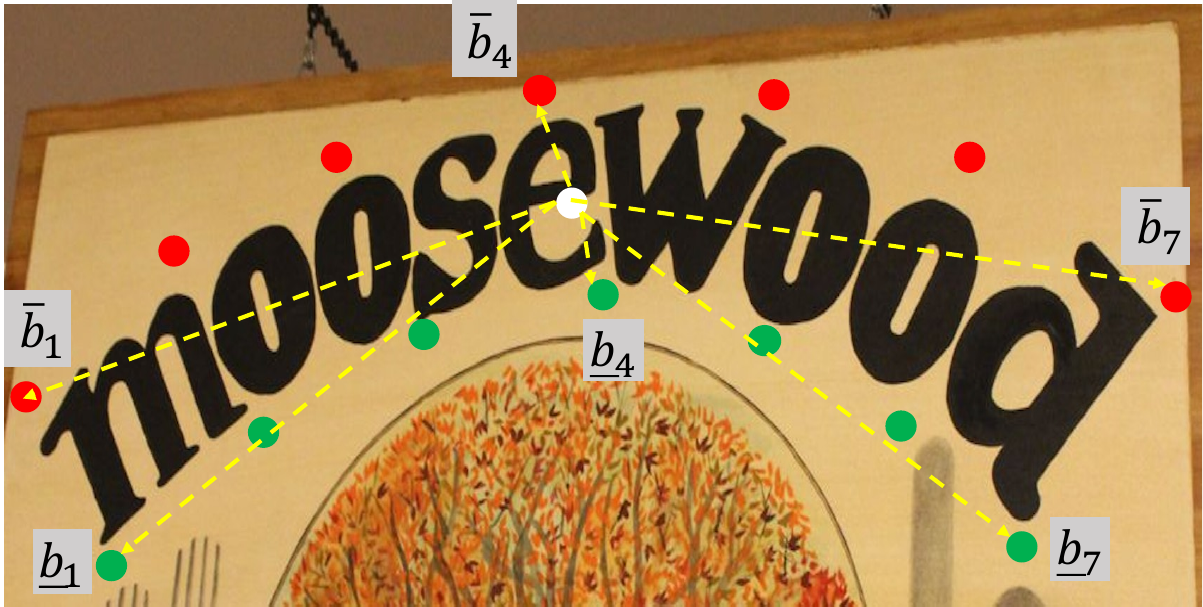}
\centering
  \caption{The representation of boundary points (points in green or red) that are regressed from the reference point (in white).}
\vspace{-2ex}  
  \label{pic:boundary}
\end{figure}

\textbf{Boundary Head} Inspired by the method~\cite{boundary}, each text-line proposal $p^l_i \in P^l$ is represented as $K$ pairs of points $\{(\overline{b}_i, \underline{b}_i) \}_{i=1}^K$, where $(\overline{b}_i, \underline{b}_i)$ denotes the $i$-th pair points localized at the up and down longer sides respectively. For each location $b_{ref} = (x, y)$ on the feature map, we can map it back onto the input image, which is near the center of the receptive field of the location. Then, we directly regress the target text proposal at the location. As shown in Fig.~\ref{pic:boundary}, $K$ pairs of coordinate offsets $(\Delta{\overline{b}}_{i},\Delta{\underline{b}}_{i}) = ((\Delta{\overline{x}}_{i},\Delta{\overline{y}}_{i}),(\Delta{\underline{x}}_{i},\Delta{\underline{y}}_{i}))$ to the boundary points are predicted at each location. Then, the predicted text proposal $\hat{p}^l$ can be calculated as 
\begin{eqnarray}
    \hat{p}^l&=&\{(b_{ref}+\Delta{\overline{b}}_{i},b_{ref}+\Delta{\underline{b}}_{i})\}^{K}_{i=1} \nonumber \\
    ~&=&\{((x + \Delta{\overline{x}}_{i}, y + \Delta{\overline{y}}_{i}),(x + \Delta{\underline{x}}_{i}, y + \Delta{\underline{y}}_{i}))\}^{K}_{i=1}
\end{eqnarray}

We employ a text-line proposal generation algorithm (TPGA) to generate the target text-line proposals $P^l$ for training. TPGA is achieved by the generating target boundary points algorithm in~\cite{boundary}, which resamples equidistantly spaced boundary points to construct the target boundary points from the up/down line of the scene text instance.

\subsection{Cross-modal Similarity Learning for the TIR Task}
The similarity task aims at measuring the similarity between query word features $\{f_{q_i}\}_{i=1}^{|Q|}$ and features $\{f_{p^l_i}\}_{i=1}^{|P^l|}$ of the text-line instances within text proposals. The feature $f_{p^l_i} \in \mathbb{R}^{T \times C}$ is extracted from feature maps by the ArbitraryRoIAlign operator in~\cite{boundary} with text-line proposal $p^l_i$.
In the training phase, the text-line proposals are not from the outputs of the text-line proposal module, as the predicted text-line proposals are not accurate enough. Instead, we form $P^l$ with the labeled bounding boxes of scene text instances. The query words $Q$ are the collection of translation of text-line instances. In other words, $q_i$ is the translation label of $p^l_i$.

The optimization target of the similarity learning task is formulated as 
\begin{equation}
  \setlength{\abovedisplayskip}{2pt}
  \label{cossim}
  sim_f(f_{q_i}, f_{p^l_j}) = sim_t(t_{q_i}, t_{p^l_j}),
\end{equation}
where $sim_f$ and $sim_t$ are functions for measuring the similarity between features and the similarity between words, respectively. The symbol $(t_{q_i}, t_{p^l_j})$ represent the corresponding translations of $(q_i, p^l_j)$, which equal to $(q_i, q_j)$ in the training stage. 

Specifically, $sim_f$ is defined as the cosine similarity between features,
\begin{equation}
  \setlength{\abovedisplayskip}{2pt}
  \label{simf}
  sim_f(f_{q_i}, f_{p^l_j}) = \frac{\tanh(v(f_{q_i}))\tanh(v(f_{p^l_j}))^{T}}{||\tanh(v(f_{q_i}))||*||\tanh(v(f_{p^l_j}))||},
\end{equation}
$v$ stands for the operator that reshapes a sequence feature of size $T \times C$ into an one-dimensional vector of size $1 \times (T*C)$.
The function $sim_t$ is the normalized edit distance between the word pairs $(t_{q_i}, t_{p^l_j})$, which is defined as Eq.~\ref{WordSim}. $Distance$ is the Levenshtein edit distance~\cite{levenshtein1966binary}, $|t_{q_i}|$ denotes the character number of $t_{q_i}$.
\begin{equation}
    \label{WordSim}
    sim_t(t_{q_i}, t_{p^l_j}) = 1-\frac{Distance(t_{q_i}, t_{p^l_j})}{\max(|t_{q_i}|, |t_{p^l_j}|)}.
\end{equation}

\subsection{MIL for the PPR Task}
Our method employs a MIL-based approach to learn the similarity between query text features and features of the partial patches. MIL-based methods require bag data for training. Therefore, before introducing the learning strategy of MIL for the PPR task, we first present details of constructing bags from text-line instances.

\subsubsection{Constructing Bags}
For each text-line proposal $p^l_i$ with its associated translation $t_{p^l_i}$, we construct a bag $P^r(p^l_i)$ and its bag label $T^r(t_{p^l_i})$. The $P^r(p^l_i)$ represents the generated proposals of partial patches on the condition of $p^l_i$. After processing all text-line proposals in $P^l$, we obtain a set of bags $P^r= \{P^r(p^l_i)\}_{i=1}^{|P^l|}$ and corresponding bag labels $T^r= \{T^r(t_{p^l_i})\}_{i=1}^{|P^l|}$, also writing as $\{p^r_i\}_{i=1}^{|P^r|}$ and $\{t_{p^r_i}\}_{i=1}^{|T^r|}$, respectively. The constructed $P^r$ and $T^r$ are used for optimizing the MIL model.

The Bag Constructing Algorithm (BCA) is designed to generate a bag $P^r(p^l_i)$ and its label $T^r(t_{p^l_i})$. 
Given the text-line proposal and its translation $(p^l_i, t_{p^l_i})$, BCA first generates a window of which width is $w*n/|t_{p^l}|$, where $w$ is the length of the longer side of $p^l_i$. Then we slide this window from the start of $p^l_i$ to its end with an interval of $w/|t_{p^l_i}|$. Meanwhile, all sub-sequences of $t_{p^l_i}$, of which length equals $n$ ($n$ characters), form the translation labels $T^r(t_{p^l}, n)$. 
{
In this way, we collect all $P^r(p^l_i, n)$ and $T^r(t_{p^l_i}, n)$ by varying $n$ in the range $[ n_{min}, n_{max}]$ to form $P^r(p^l_i)$ and $T^r(t_{p^l_i})$. 
To iterate over all possible sizes of proposals, we set $n_{min}$ and $n_{max}$ as 2 and $|t_{p^l}|$, respectively.
}
More detailed implementations are given in Alg.~\ref{alg:ppga}. 

\begin{figure}[t]
    \includegraphics[width=1.0\linewidth]{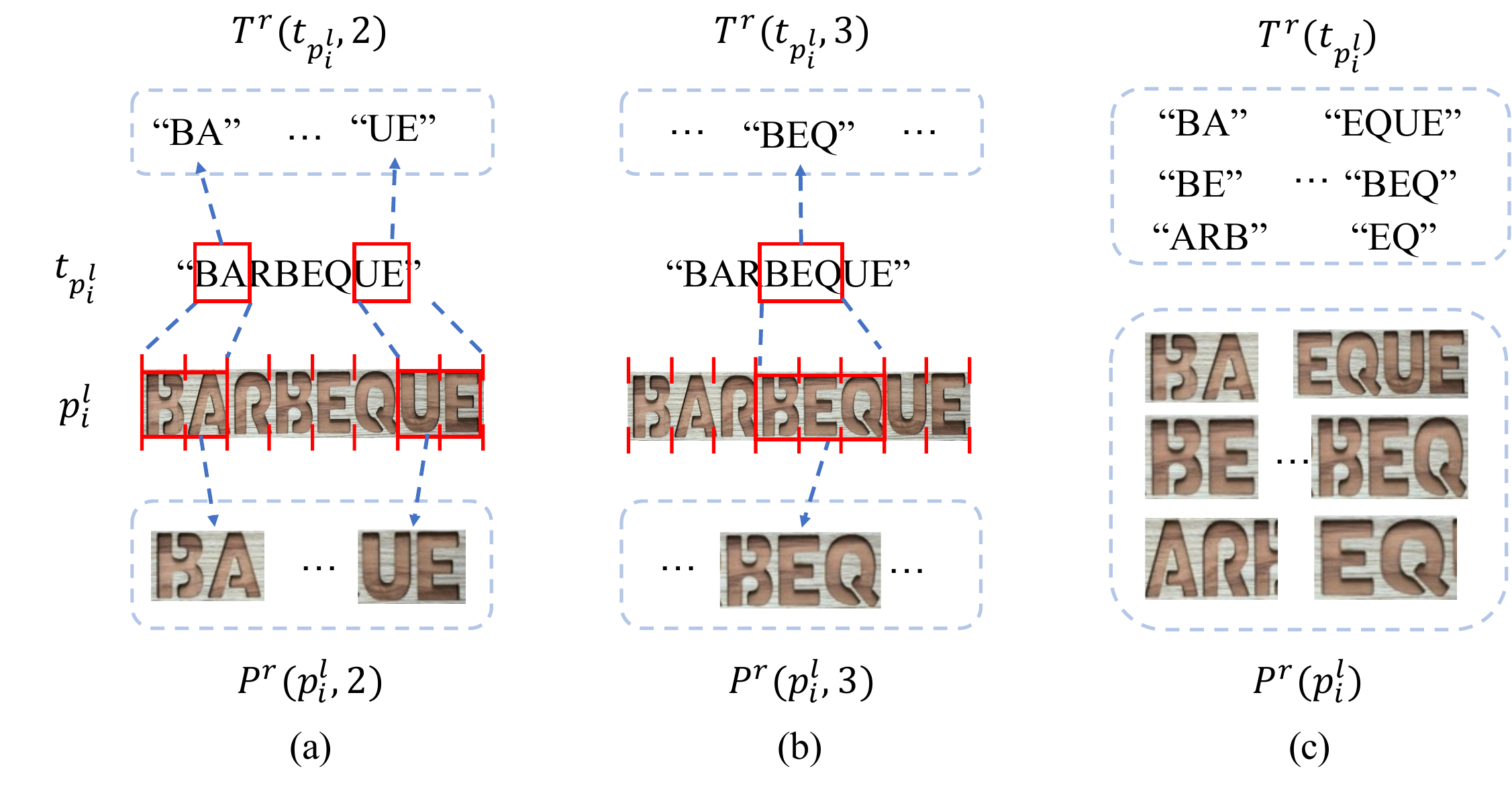}
\centering
\vspace{-5ex}
  \caption{An example of constructing a bag (c) from the labeled text-line instance.}
\vspace{-1ex}  
  \label{mil_case}
\end{figure}

As in Fig.~\ref{mil_case}, (a) and (b) illustrate the process of generating partial text proposals and their associated translations when $n$ equals 2 and 3, respectively. In Fig.~\ref{mil_case} (c), we collect all generated partial text proposals to construct $P^r(p^l_i)$ and $T^r(t_{p^l_i})$. 
{
Since we can use boundary points to rectify arbitrary-shaped text instances into horizontal ones, text instances of any shape can be processed using the same pipeline.
} 

\begin{algorithm}
    \caption{Bag Constructing Algorithm}
    \label{alg:ppga}
    \begin{algorithmic}[1]
        \REQUIRE ~~ \\
        Text-line proposal: $p = \{(\overline{b}_i, \underline{b}_i) \}_{i=1}^{|p|}$.\\
        Text-line translation: $t = \{ c_i \}_{i=1}^{|t|}$.\\
        The window length range : $[ n_{min}, n_{max}]$. 
        \ENSURE ~~ \\
        Generated partial proposal set : $P^r(p) = \{\} $.\\
        Translations of $P^r(p)$: $T^r(t) = \{\} $.\\
        \STATE $//$ TPGA is introduced in Section~\ref{sec:bpm}.
        \STATE $\tilde{p}$ = TPGA($p$, $|t|+1$); $// \tilde{p} = \{(\overline{\tilde{b}}_i, \underline{\tilde{b}}_i) \}_{i=1}^{|t|+1}$
        \FOR{$n=n_{min}; n <= n_{max}; n++$}
        \STATE $P^r(p, n), T^r(t, n) = \{\}, \{\} $;
            \FOR{$i=1; i < |t| - n; i++$}
            \STATE $\tilde{p}_i$ = TPGA( $\{(\overline{\tilde{b}}_j, \underline{\tilde{b}}_j) \}_{j=i}^{i+n}$, $|p|$);
            \STATE $P^r(p, n) = P^r(p, n) \cup \{ \tilde{p}_i \} $;
            \STATE $T^r(p, n) = T^r(p, n) \cup \{ \{ c_j \}_{j=i}^{i+n} \} $;
            \ENDFOR
        \STATE $P^r(p) = P^r(p) \cup P^r(p, n) $;
        \STATE $T^r(t) = T^r(t) \cup T^r(t, n) $;
        \ENDFOR
        \RETURN $P^r(p)$ and $T^r(t)$;
    \end{algorithmic}
\end{algorithm}

\subsubsection{Learning Strategy of MIL}

The MIL network aims at learning to measure the similarity between a query word feature and feature of a partial patch, by using the constructed bags $(P^r=\{p^r_i\}_{i=1}^{|P^r|}, T^r=\{t_{p^r_i}\}_{i=1}^{|T^r|})$. A feasible solution is to learn such similarity with the cross-modal similarity learning strategy by treating the constructed label $t_{p^r_i}$ as the ground-truth translation of each proposal $p^r_i$. Yet, the BCA is based on the assumption that each character of $p^l$ occupies equal spatial space of a text-line instance. It is relatively strong and hardly holds especially when the shape of a text-line instance is not horizontal. This leads that the associated pseudo translation $t_{p^r_i}$ of $p^r_i$ is not enough accurate and can not be used to optimize the similarity between $f_{p^r_i}$ and $f_{t_{p^r_i}}$ by Eqn.~\ref{cossim}.

For the query text $q_j$, we know whether $t_{q_j}$ is contained in $t_{p^l_i}$. Therefore, we can know whether the bag $P^r(p^l_i)$ contains a patch translated as $t_{q_j}$ as long as we sample enough patches from $p^l_i$. According to a typical MIL method, we can formulate the optimization target as
\begin{equation}
  \setlength{\abovedisplayskip}{2pt}
  \label{mil}
  sim_f(f_{q_j}, f_{P^r(p^l_i)}) = \mathbb{I}(t_{q_j} \exists T^r(t_{p^l_i})),
\end{equation}
where the function $\mathbb{I}(\Psi)$ equals to 1 if the condition $\Psi$ holds otherwise 0. The condition $t_{q_j} \exists T^r(t_{p^l_i})$ denotes $t_{q_j}$ is contained in $T^r(t_{p^l_i})$. The $f_{P^r(p^l_i)}$ is an aggregated features of all partial patches in $P^r(p^l_i)$. The aggregation function in MIL usually is a max-pooling or average-pooling. We define the aggregation function as follow,
\begin{equation}
	f_{P^r(p^l_i)} = f_{p^r_{\theta}} = \mathop{\arg\max}_{f_x} \ \ sim_f(f_{q_j}, f_x), x \in P^r(p^l_i).
	\label{aggregation}
\end{equation}
This function means that we represent $f_{P^r(p^l_i)}$ as a patch feature $f_{p^r_{\theta}}$ that has the maximum similarity to $f_{q_j}$. 

\subsection{The Proposed RankMIL for the PPR Task}
Ideally, for the conventional MIL, if query text $q_j$ is contained in the constructed bag label $T^r(t_{p^l_i})$, the translation of the selected patch $p^r_{\theta}$ from bag should equal the given query text. However, this hypothesis is difficult to hold, mainly attributing to two aspects. First, a bag label contains the given query text, while the bag does not include the partial patch translated as this query text, producing false positive bag. Second, although the bag contains the partial patch translated as query text, the measurement errors from the aggregation function in Eqn.~\ref{aggregation} make the selected patch not the right one. The two cases can largely mislead the optimization by Eqn.~\ref{mil}.

We observe that the bag $P^r(p^l_i)$ always contains a patch that is more similar to the given query text than the text-line instance $p^l_i$. Hence, we turn the optimization target of MIL from the binary classification to a type of ranking loss. The proposed RankMIL optimizes the similarity between query text and the selected patch larger than the similarity between this query text and the text-line instance a margin. In order to eliminate the noisy samples brought by the measurement errors of the aggregation function, we abandon these selected patches $p^r_{\theta}$ of which similarities are less than the similarity between query text and the text-line instance. The optimization target of RankMIL is formulated as follows,

\begin{equation}
\begin{aligned}
	loss(f_{q_j}, f_{p^r_{\theta}}) &= \max(0, -\mathbb{I}(t_{q_j} \exists T^r(t_{p^l_i}) \land \Delta{s}>0)(\Delta{s}-m))\\
	\Delta{s} &= sim_f(f_{q_j}, f_{p^r_{\theta}}) -sim_f(f_{q_j},f_{p^l_i}),
\end{aligned}
\label{loss:seq_mil}
\end{equation}
where $f_{p^r_{\theta}}$ is defined as Eqn.~\ref{aggregation}, and $m$ is a constant number.

\begin{figure}[t]
    \includegraphics[width=0.98\linewidth]{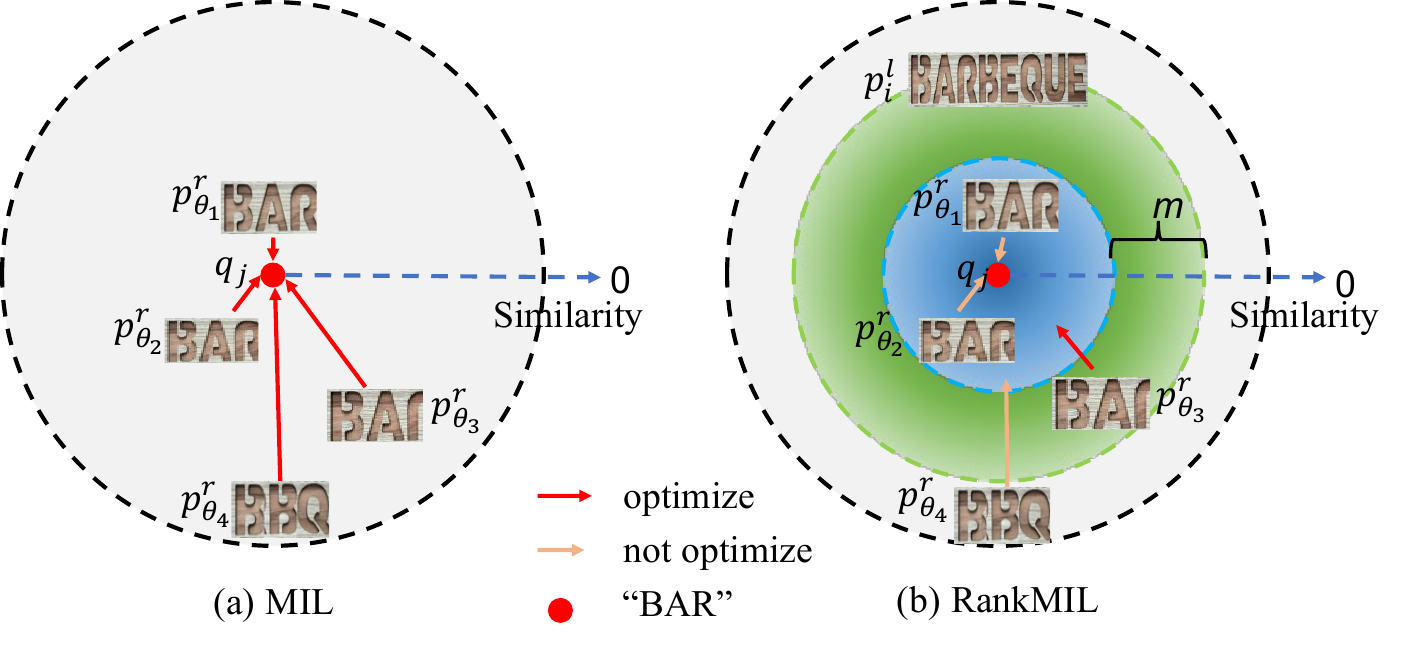}
\centering
\vspace{-1.5ex}
  \caption{The optimization target comparison between conventional MIL (a) and the RankMIL (b). The red point is a query text $q_j$. The red arrow and yellow arrow indicate training and abandoning this sample, respectively. In (b), from the outermost to the inner, the similarity value at the three dash circles equals to 0, the similarity between $q_j$ and $p^l_i$, and the similarity with a margin $m$ over the similarity between $q_j$ and $p^l$.} 
\vspace{-2ex}  
  \label{pic:seq_mil}
\end{figure}

As illustrated in Fig.~\ref{pic:seq_mil}, we compare the conventional MIL and the RankMIL under four types of selected partial patches $p^r_{\theta_1},p^r_{\theta_2},p^r_{\theta_3}, p^r_{\theta_4}$. The conventional MIL optimizes the similarity between each $p^r_{\theta}$ and $q_j$ to 1 whether this partial patch is a noisy sample ($p^r_{\theta_4}$) or not. Among the four cases, the RankMIL only optimize similarity of patch $p^r_{\theta_3}$ with the query text to the range {$[sim_f(f_{q_j},f_{p^l_i})+m, 1]$}. For $p^r_{\theta_4}$ which is less similar to $q_j$ than $p^l_i$, we think it is a noisy sample caused by the prediction errors of the aggregation function. Besides, the RankMIL does not force the similarities of those enough similar instances ($p^r_{\theta_1}$ and $p^r_{\theta_2}$) with $q_j$ to 1, as they might contain some background. Not continuing to optimize them can reduce disturbances to the similarity measurement.

\subsection{Dynamic Partial Match Algorithm}

During the inference phase, the network can detect proposals for text-line instances but not for partial patches. To address this issue, a feasible solution is to sample partial patches from detected text-line proposals using the BCA while assuming that each text-line instance has a fixed number of characters. However, this method only provides continuous partial patches, making it challenging to retrieve non-continuous ones. Moreover, calculating similarities between query text and all partial patches is a time-consuming process.
To overcome these limitations, the DPMA directly searches for the feature of the most similar partial patch with the given query text, without the need to enumerate all partial patches.

\begin{figure}[t]
    \includegraphics[width=0.98\linewidth]{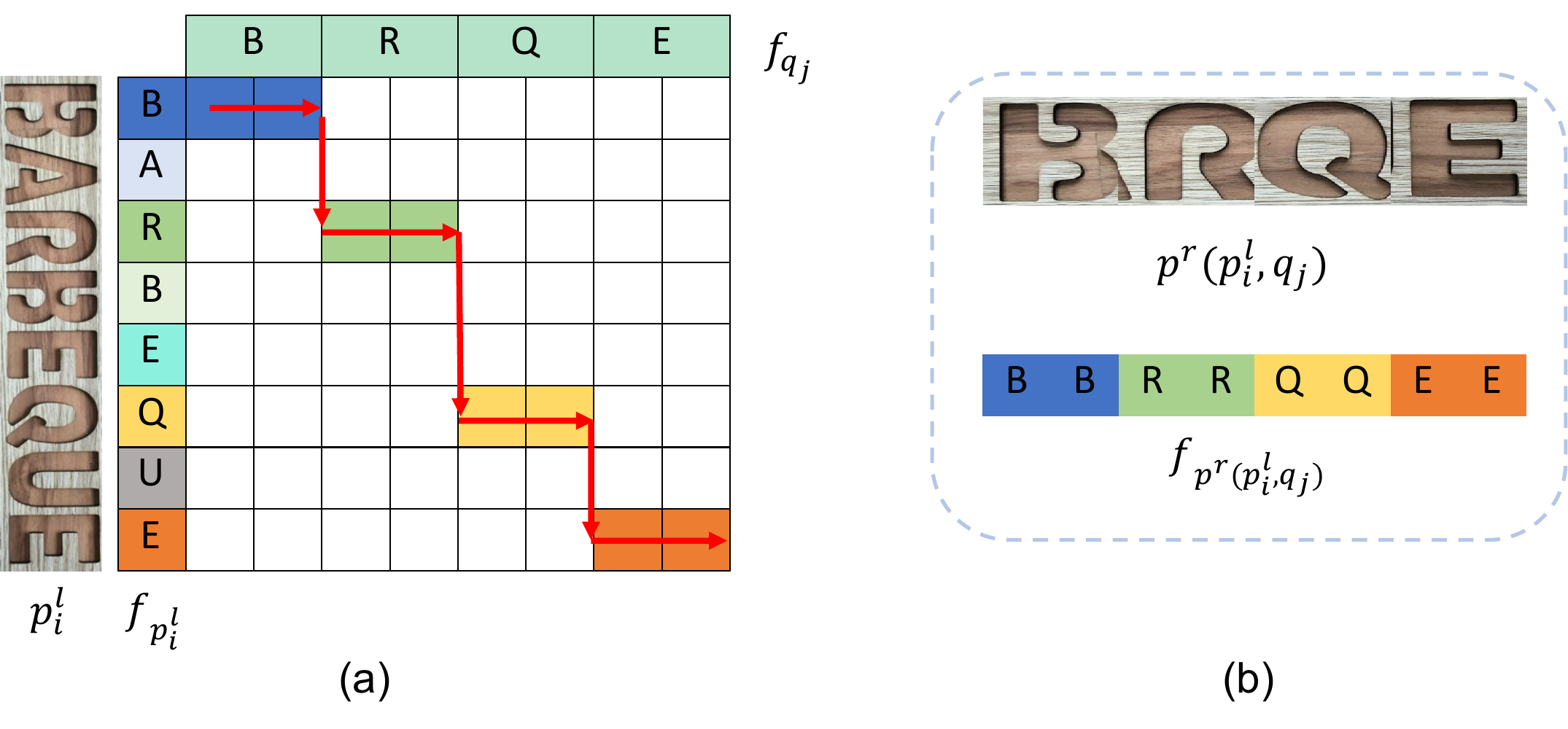}
\centering
\vspace{-2ex}
  \caption{The illustration of the DPMA. (a): Features $f_{q_j}$ and $f_{p^l_i}$ consist the two-demensional search grid with size $T \times T$ (we set $T$ to 8 for visualization convenience). The path denoted as red arrow line is the searched optimal path. (b) The searched partial patch within $p^r(p^l_i,q_j)$ and the extracted features.} 
\vspace{-2ex}  
  \label{pic:dmpp}
\end{figure}

Specifically, the input sequence features, $f_{p^l_i} \in \mathbb{R}^{T \times C}$ and $f_{q_j} \in \mathbb{R}^{T \times C}$, compose a two-dimensional search grid with size $T \times T$, as shown in Fig.~\ref{pic:dmpp} (a). The proposed DPMA utilizes dynamic programming to search for the optimal path. The search process is guided by three rules. First, the search range spans from the first to the last column on the grid. Second, since both the scene text and query text are types of ordered time series, the search proceeds in a one-way direction, with the path walking one step to the right and at least zero steps down at each step. Finally, the objective of the search is to maximize the cumulative similarity score along the path.
The state transition function of this dynamic programming is as follows,
\begin{equation}
S_{x,y}=\left\{
                    \begin{array}{ll}
                    s_{x,y} & y=0\\
                    S_{x,y-1}+s_{x,y} & x=0,y\neq 0\\
                    max(\{S_{k,y-1}\}_{k=0}^x)+s_{x,y} & x\neq0,y\neq0
                    \end{array}, \right.
\end{equation}
where $S_{x,y}$ is the cumulative similarity of the optimal path when walking to $(x,y)$ in the grid. The $s_{x,y}$ is the similarity in the current cell of the grid, which is calculated by Eqn.~\ref{simf} with $f_{p^l_i}[x] \in \mathbb{R}^{1 \times C}$ and $f_{q_j}[y] \in \mathbb{R}^{1 \times C}$ as inputs.

After obtaining the optimal path, we in order sample the feature located on the path where the optimal path is cast to the $f_{p^l}$-axis, as illustrated in Fig.~\ref{pic:dmpp} (b). All sampled features are concatenated as new features $f_{p^r(p^l_i, q_j)} \in \mathbb{R}^{T \times C}$. The $p^r(p^l_i, q_j)$ indicates the proposal of partial patch searched from the text-line proposal $p^l_i$ with the query text $q_j$. The similarity between the searched partial patch and the query text equals the similarity between $f_{p^r(p^l_i, q_j)}$ and $f_{q_j}$, which is calculated by Eqn.~\ref{simf}.

\begin{algorithm*}[t]
\caption{Ranking images in a gallery by the DPMA (left) or constructing bags (right).} 
\begin{minipage}{.5\textwidth}
    \begin{algorithmic}[1]
    \label{alg:inference}
    \REQUIRE ~~\\
    The image gallery, $G = \{g_i\}_{i=1}^{|G|}$;\\
    The query text, $q$;
    \ENSURE ~~\\
    The ranking scores, $S = \{s_i\}_{i=1}^{|G|}$;
    \STATE Extract features of query text, $f_q$; \label{ code:fram:extract}
    \STATE Input an image, $g_i$;\label{code:start}
    \STATE Detect text-line proposals, $P^l=\{p^l_i\}_{i=1}^{|P^l|}$;\\
    \STATE Extract features within $P^l$, $\{f_{p^l_i}\}_{i=1}^{|P^l|}$; \label{ code:fram:extract}
    \STATE Search features of partial instances within $p^l_i$ by the DPMA,\\ $\{f_{p^r(p^l_i, q)}\}_{i=1}^{|P^l|}$;\\
    \STATE Collect features of $P^l$ and partial instances,\\
    $F = \{f_{p^r(p^l_i, q)}\}_{i=1}^{|P^l|} \cup \{f_{p^l_i}\}_{i=1}^{|P^l|}$;\\
    \STATE Calculate the score of $g_i$,\\ $s_i = \mathop{\max}_j(\{sim_f(f_q,f_j), f_j\in F\})$;\label{code:end}
    \STATE Calculate each $s_i$ by looping from step~\ref{code:start} to step~\ref{code:end}. 
    \RETURN $S$;
    \end{algorithmic}
\end{minipage}
\begin{minipage}{.5\textwidth}
    \begin{algorithmic}[1]
    \REQUIRE ~~\\
    The image gallery, $G = \{g_i\}_{i=1}^{|G|}$;\\
    The query text, $q$;
    \ENSURE ~~\\
    The ranking scores, $S = \{s_i\}_{i=1}^{|G|}$;
    \STATE Extract features of query text, $f_q$; \label{ code:fram:extract}
    \STATE Input an image, $g_i$;\label{code:start}
    \STATE Detect text-line proposals, $P^l=\{p^l_i\}_{i=1}^{|P^l|}$;\\
    \STATE Construct bags within each $p^l_i$ by the BCA,\\ $P^r = \{P^r(p^l_i)\}_{i=1}^{|P^l|}$;\\
    \STATE Collect proposals from $P^l$ and $P^r$,\\
    $P = \{p^l_i\}_{i=1}^{|P^l|} \cup \{P^r(p^l_i)\}_{i=1}^{|P^l|} $;\\
    \STATE Extract features within $P$, $F$; \label{ code:fram:extract}
    \STATE Calculate the score of $g_i$,\\ $s_i = \mathop{\max}_j(\{sim_f(f_q,f_j), f_j\in F\})$;\label{code:end}
    \STATE Calculate each $s_i$ by looping from step~\ref{code:start} to step~\ref{code:end}. 
    \RETURN $S$;
    \end{algorithmic}
\end{minipage}
\end{algorithm*}

In Alg.~\ref{alg:inference}, we compare the inference pipeline of using the DPMA and constructing bags. Our method achieves TIR and PPR tasks with the same trained model and inference pipeline.
In the inference phase, given query text and an image gallery $G = \{g_i\}_{i=1}^{|G|}$, our approach outputs for each image a ranking score indicating the probability that this image contains the given query text. Compared to retrieving text by constructing bags, the way with the DPMA not requires generating bags, greatly saving the inference time.
Note that we define the probability of an image containing the query text as the highest score among all text regions in that image. In fact, all target text regions within the image can be identified using a similarity threshold.

\subsection{Optimization}\label{sec:optimization}
The objective function consists of three parts, which is defined as follows;
\begin{equation}
    L = L_{det} + L_{cms} +L_{mil},
\end{equation}
where $L_{det}$ is the detection loss, which comes from three heads of the text-line proposal module. The loss functions of the classification head and the centerness head are identical to these in~\cite{tian2019fcos}. The objective function of the boundary head is the ones in~\cite{boundary}. $L_{mil}$ is the loss function of the RankMIL. For each query text, we optimize its similarity with a bag by Eqn.~\ref{loss:seq_mil}. 

$L_{cms}$ aims to optimize the cross-modal similarity learning task, which is formulated as
\begin{equation}
    L_{cms} = L_{sim}(Q,P^l) + L_{sim}(P^l,P^l) + L_{sim}(Q,Q).
\end{equation}
$L_{sim}(Q,P^l)$ is the loss function of measuring similarity between all query text and all text-line instances in $P^l$. {
Since query text and text-line instances are different modal data, directly minimizing $L_{sim}(Q, P^l)$ is challenging. Therefore, to facilitate cross-modal similarity learning, we also learn the similarity within the same modality data, including the query text data and the text-line instance data.} The three kinds of loss functions have the same formulation except for the inputs.
Taking $L_{sim}(Q,P^l)$ as example, the function is formulated as
\begin{equation}
    L_{sim}(Q,P^l) = \sum_{i}^{|Q|}\mathop{\max}_j(|sim_f(f_{q_i}, f_{p^l_j})-sim_t(t_{q_i}, t_{p^l_j})|).
\end{equation}
Inspired by the approach~\cite{conf/cvpr/ShrivastavaGG16}, we apply hard example mining technique to focus on optimizing the similarity between similar samples by the operator $\max()$.

\section{Experiments}
First, we introduce the datasets used in our experiments and the newly created dataset. Second, the implementation details are given. 
Third, we verify the ablation studies on our proposed RankMIL and DPMA. 
Then, we evaluate our method and make comparisons with the state-of-the-art scene text retrieval/spotting approaches on Chinese/English datasets.
Last, we compare with other methods on online and offline inference speeds.
{The evaluation metric mean average precision (mAP) in our experiments is the same as the metric used in~\cite{GomezECCV20118YOLO+PHOC}.}

\subsection{Datasets}
The datasets used in our experiments are divided to two groups: English datasets and Chinese datasets. The statistics of each dataset are listed in Tab.~\ref{dataset_summary}. For the PPR task, we generate query words by cutting each query word of the TIR task into sub-sequences by the thulac tools\footnote{https://github.com/thunlp/THULAC-Python.git}~\cite{journals/coling/LiS09}. The query text of the PPR task is used for evaluating the performance of retrieving continuous partial patches (CPP) and non-continuous partial patches (NCPP).

\begin{table}[t]
  \caption{
  The summary of evaluation datasets. The TIR Num. and PPR Num. mean the number of query words for evaluations on the TIR task and the PPR task, respectively. (CPP, NCPP) denotes that a query word appears in a continuous or non-continuous partial patch.}
  \vspace{-2ex}
  \centering
  \begin{tabular}{
    m{.18\columnwidth}|m{.1\columnwidth}<{\centering}
    m{.12\columnwidth}<{\centering}| m{.08\columnwidth}<{\centering}
    m{.08\columnwidth}<{\centering} m{.18\columnwidth}<{\centering}}
  \Xhline{1.0pt}
  Dataset& Script& Text shape& Image Num.& TIR Num.& PPR Num. (CPP, NCPP)\\
  \hline  
  STR~\cite{MishraAJ13}& English& Horizontal& 10000& 50& 48 (26, 22)\\
  CTR~\cite{COCOText}& English& Oriented& 7196& 500& 98 (37, 61)\\
  ArT~\cite{conf/icdar/ChngDLKCJLSNLNF19}& English& Curved& 5603& 1022& 166 (97, 69)\\
  \cline{1-6}
  {CSVTRv2}& Chinese& Oriented& {3400}& 53& 103 (65, 38)\\
  ReCTS~\cite{conf/icdar/ZhangYBSKLJZJSL19}& Chinese& Oriented& 2000& 584& 483 (406, 77)\\
  LSVT~\cite{conf/icdar/SunKCJNCLLNHDL19}& Chinese& Oriented& 3000& 541& 351 (282, 69)\\
  \Xhline{1.0pt}
  \end{tabular}
  \label{dataset_summary}
  \vspace{-2ex}
\end{table}

\textit{IIIT Scene Text Retrieval dataset} (STR)~\cite{MishraAJ13} consists of 50 query words and 10,000 images. We generate 48 query words for the PPR task. It is a challenging dataset due to the variation of fonts, styles and view points.

\textit{Coco-Text Retrieval dataset} (CTR) is a subset of Coco-Text~\cite{COCOText}. We select 500 annotated words and 98 sub-sequences from Coco-Text as queries for the TIR task and the PPR task, respectively. Then, 7196 images in Coco-Text containing such query words are used to form this dataset.

\textit{Arbitrary-Shaped Text dataset} (ArT)~\cite{conf/icdar/ChngDLKCJLSNLNF19} is currently the largest dataset for arbitrarily shaped scene text. It is the combination of the Total-Text~\cite{TT}, SCUT-CTW1500~\cite{journals/pr/LiuJZLZ19}, and the newly collected images.
The new images also contains at least one arbitrarily-shaped
text per image. The ArT dataset is split to a training set with
5,603 images and 4,563 for testing set. The 5603 images and the top 1022 most frequent words and 166 sub-sequences are selected to form testing set for retrieval task.

\textit{SynthText-900k dataset}~\cite{GomezECCV20118YOLO+PHOC} is composed of about 925,000 synthetic English text images, generated by a synthetic engine~\cite{GuptaVZ16} with slight modifications.

\textit{Multi-lingual Scene Text 5k dataset} (MLT-5k) is a subset of MLT~\cite{MLT19}, which consists of about 5000 images containing text in English.

\textit{Chinese Street View Text Retrieval dataset} ({CSVTRv2}) consists of 53 pre-defined query words in Chinese and 3400 Chinese scene text images collected from the Google image search engine. We generate and select 103 sub-sequences for the PPR task. Each image is annotated with its corresponding query word among the 53 pre-defined Chinese query words. Most query words are the names of business places.

\textit{Reading Chinese Text on Signboard dataset} (ReCTS)~\cite{conf/icdar/ZhangYBSKLJZJSL19} contains 25k annotated signboard images that mainly contains text of the shop signs. There are 20k images in training set, and the rest are for testing set. As the annotations in the testing set are not provided, we split the training set into two parts to form  training set (18k images) and testing set (2k images) for the retrieval task. We select the top 584 most frequent words and 483 sub-sequences as query text.

\textit{Large-scale Street View Text dataset} (LSVT)~\cite{conf/icdar/SunKCJNCLLNHDL19} provides an unprecedentedly large number of text from street view. It provides total 450k images with rich information of the real scene, among which 50k images are annotated in full annotations (30k images for training and the rest 20k images for testing). We select the top 541 most frequent words and 351 sub-sequences as query text.

\textit{SynthText-100k dataset}~\cite{journals/corr/abs-2105-03620} consists of about 100k synthetic images where the text instances are in Chinese. This dataset is generated by a synthetic engine~\cite{GuptaVZ16} with slight modifications.

In the experiments of English text retrieval, STR, CTR, and ArT are the testing datasets, while SynthText-900k and MLT-5k are only used for training the proposed model. For the experiments of Chinese text retrieval, {CSVTRv2}, ReCTS, and LSVT are the testing datasets, while SynthText-100k and the training set of both ReCTS and LSVT are employed for training the model.

\begin{table}[t]
  \caption{
  TIR results with different methods of PHOC, End-to-end recognition (E2E) and the CMSL. (Metric: mAP)}
  \vspace{-2ex}
  \centering
  \begin{tabular}{
    m{.22\columnwidth}|
    m{.1\columnwidth}<{\centering} m{.1\columnwidth}<{\centering}m{.1\columnwidth}<{\centering}|
    m{.1\columnwidth}<{\centering}m{.1\columnwidth}<{\centering} }
  \Xhline{1.0pt}
  \multirow{2}*{Method}& \multicolumn{3}{c|}{Chinese} & \multicolumn{2}{c}{English}\\
  \cline{2-6}
  & {CSVTRv2}& ReCTS& LSVT& STR& ArT\\
  \hline  
  Baseline+PHOC& -& -& -& 72.41& 68.26\\
 Baseline+E2E& 86.06& 70.48& 55.26& 76.39& 73.12\\
 Baseline+CMSL& \textbf{89.45}& \textbf{73.00}& \textbf{60.50}& \textbf{80.15}& \textbf{78.04}\\
  \Xhline{1.0pt}
  \end{tabular}
  \label{abla:cmsl}
  \vspace{-2ex}
\end{table}

\begin{table}[t]
\centering
\vspace{2ex}
\caption{{The impact of hyperparamerter $K$ on the TIR task. Experiments with rectangle mean text proposals are formulated as horizontal boxes.}}
\begin{tabular}{l|c|cccc}
\Xhline{1.0pt}
&\multirow{2}*{Rectangle}& \multicolumn{4}{c}{Boundary ($K$)}\\
\cline{3-6}
&  & 3 &5    & 7  & 9\\ \hline
 ArT& 70.00 & 72.71& 75.32& 77.94& 77.62\\
 LSVT& 58.31& 59.12& 60.23& 60.50& 60.12\\
 \Xhline{1.0pt}
\end{tabular}
\vspace{-2ex}
\label{abl_k}
\end{table}

\begin{figure}[t]
    \includegraphics[width=0.98\linewidth]{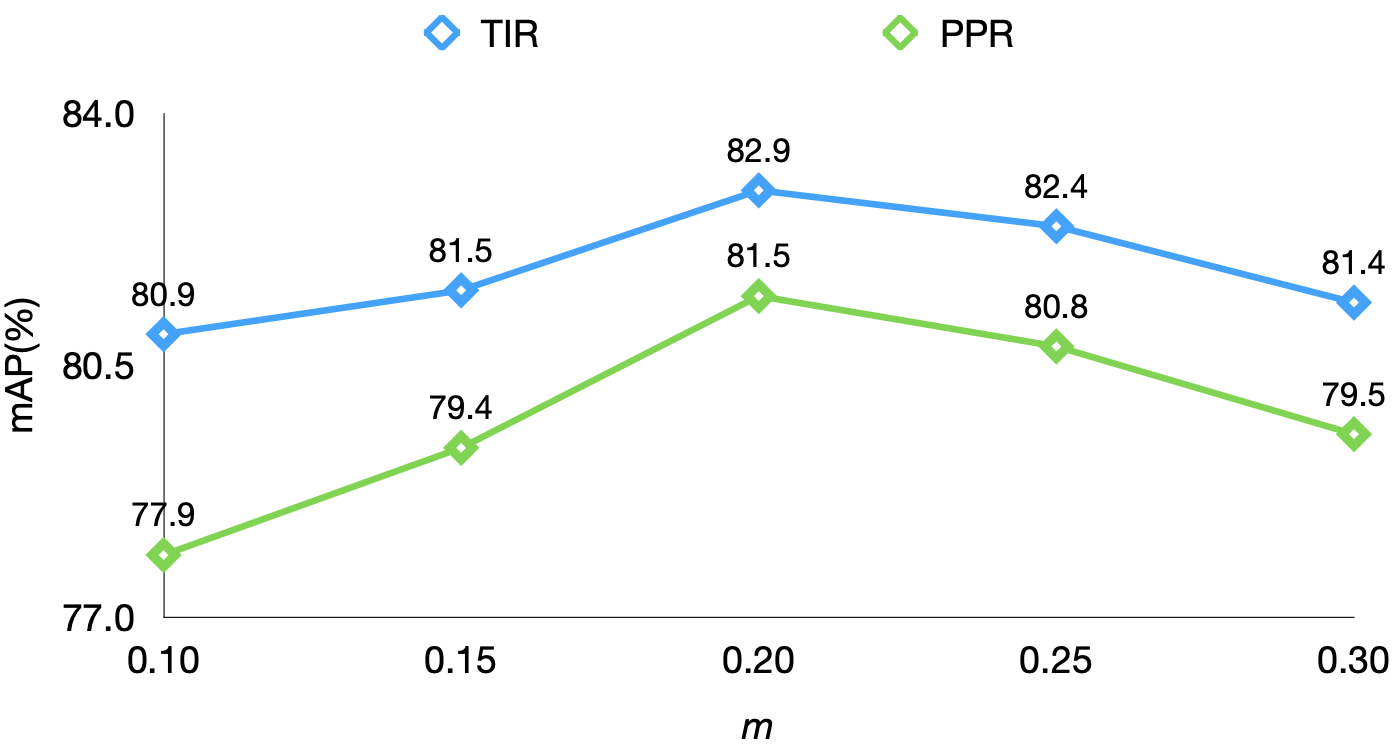}
\centering
\vspace{-1ex} 
  \caption{The influence of the parameter $m$ in the RankMIL to the average performances of the TIR and PPR tasks.}
\vspace{-2ex}  
  \label{fig:margin}
\end{figure}

\begin{table*}[t]
  \caption{
  Comparisons of different learning strategies on TIR and PPR tasks. (Metric: mAP)}
  \vspace{-2ex}
  \centering
  \begin{tabular}{
    m{.15\columnwidth}<{\centering}|
    m{.04\columnwidth}<{\centering}m{.04\columnwidth}<{\centering}|
    m{.07\columnwidth}<{\centering} m{.07\columnwidth}<{\centering}|
    m{.07\columnwidth}<{\centering} m{.07\columnwidth}<{\centering}|
    m{.07\columnwidth}<{\centering} m{.07\columnwidth}<{\centering}|
    m{.07\columnwidth}<{\centering} m{.07\columnwidth}<{\centering}|
    m{.07\columnwidth}<{\centering} m{.07\columnwidth}<{\centering}|
    m{.07\columnwidth}<{\centering} m{.07\columnwidth}<{\centering}|
    m{.07\columnwidth}<{\centering} m{.07\columnwidth}<{\centering}}
  \Xhline{1.0pt}
  \multirow{2}*{\makecell[l]{Learning \\strategy}}&\multicolumn{2}{c|}{Bag data}  & \multicolumn{2}{c|}{{CSVTRv2}} & \multicolumn{2}{c|}{ReCTS}& \multicolumn{2}{c|}{LSVT}&\multicolumn{2}{c|}{Avg. (Chinese)} 
  & \multicolumn{2}{c|}{STR} & \multicolumn{2}{c|}{ArT}&\multicolumn{2}{c}{Avg. (English)}\\
  \cline{2-17}
  & $T^r$& $P^r$&TIR& PPR& TIR& PPR& TIR& PPR& TIR& PPR& TIR& PPR& TIR& PPR& TIR& PPR\\
  \hline  
CMSL-A
  &       &       &  88.63& 68.70& 79.01& 75.96& 68.82& 66.05& 74.76& 70.45&  78.49& 56.36& 74.12& 47.52& 74.32& 49.50\\
  CMSL-B
  &$\surd$&       &  90.63& 70.66& 78.52& 76.39& 67.65& 66.18& 74.07& 71.94&  78.03& 57.76& 73.06& 48.87& 73.29& 50.86\\
  CMSL-C
  &$\surd$&$\surd$&  89.17& 70.60& 78.17& 76.04& 67.65& 61.31& 73.83& 69.92&  77.57& 56.39& 72.86& 47.25& 73.08& 49.30\\
  \hline
  MIL
  &$\surd$&$\surd$&  \textbf{90.81}& 71.81& 80.78& 77.45& 72.65& 68.39& 77.50& 73.44&  79.78& 58.67& 76.45& 50.12& 76.61& 52.04\\
  RankMIL
  &$\surd$&$\surd$&  90.50& \textbf{75.74}& \textbf{82.93}& \textbf{81.64}& \textbf{73.86}& \textbf{71.90}& \textbf{79.11}& \textbf{77.34}&  \textbf{81.02}& \textbf{62.12}& \textbf{78.18}& \textbf{53.44}& \textbf{78.31}& \textbf{55.39}\\
  \Xhline{1.0pt}
  \end{tabular}
  \label{ablation:seq_mil}
  \vspace{-2ex}
\end{table*}

\subsection{Implementation Details}
\textbf{English Text Retrieval} The whole training process includes two steps: pre-trained on SynthText-900k and fine-tuned on MLT-5k. In the pre-training stage, we set the mini-batch to 64, and input images are resized to 640 $\times$ 640. In the fine-tuning stage, data augmentation is applied. Specifically, the longer sides of images are randomly resized from 640 pixels to 1920 pixels. Next, images are randomly rotated in a certain angle range of [$-15^{\circ}, 15^{\circ}$]. Then, the heights of images are rescaled with a ratio from 0.8 to 1.2 while keeping their widths unchanged. Finally, 640 $\times$ 640 image patches are randomly cropped from the images. The mini-batch of images is kept to 16.

We optimize our model using SGD with a weight decay of 0.0001 and a momentum of 0.9. In the pre-training stage, we train our model for 90k iterations, with an initial learning rate of 0.01. Then the learning rate is decayed to a tenth at the 30k$^{th}$ iteration and the 60k$^{th}$ iteration, respectively. In the fine-tuning stage, the initial learning rate is set to 0.001, and then is decreased to 0.0001 at the 40k$^{th}$ iteration. The fine-tuning process is terminated at the 80k$^{th}$ iteration. For testing, the longer sides of input images are resized to 960 while the aspect ratios are kept unchanged.

\textbf{Chinese Text Retrieval} The training process of the Chinese text retrieval experiments is similar to the ones of English text retrieval, except the pre-training dataset (SynthText-100k) and the  fine-tuning datasets (the training set of both ReCTS and LSVT). Besides, in the fine-tuning stage, the initial learning rate is set to 0.001, and then is decreased to 0.0001 at the 60k$^{th}$ iteration. The fine-tuning process is terminated at the 100k$^{th}$ iteration.

All experiments are conducted on a regular workstation with NVIDIA Titan Xp GPUs with Pytorch. The model is trained in parallel on four GPUs and tested on a single GPU.

\begin{table*}[t]
  \caption{
  The influence of the DPMA on TIR and PPR tasks. (Metric: mAP)}
  \vspace{-2ex}
  \centering
  \begin{tabular}{
    m{.24\columnwidth}|
    m{.07\columnwidth}<{\centering} m{.07\columnwidth}<{\centering}|
    m{.07\columnwidth}<{\centering} m{.07\columnwidth}<{\centering}|
    m{.07\columnwidth}<{\centering} m{.07\columnwidth}<{\centering}|
    m{.07\columnwidth}<{\centering} m{.07\columnwidth}<{\centering}|
    m{.07\columnwidth}<{\centering} m{.07\columnwidth}<{\centering}|
    m{.07\columnwidth}<{\centering} m{.07\columnwidth}<{\centering}|
    m{.07\columnwidth}<{\centering} m{.07\columnwidth}<{\centering}}
  \Xhline{1.0pt}
& \multicolumn{2}{c|}{{CSVTRv2}} & \multicolumn{2}{c|}{ReCTS}& \multicolumn{2}{c|}{LSVT}& \multicolumn{2}{c|}{Avg. (Chinese)}& \multicolumn{2}{c|}{STR}& \multicolumn{2}{c|}{ArT}& \multicolumn{2}{c}{Avg. (English)} \\
  \cline{2-15}
  & TIR& PPR& TIR& PPR& TIR& PPR& TIR& PPR& TIR& PPR& TIR& PPR& TIR& PPR\\
  \hline  
  Baseline & 89.45& 45.42& 73.00& 46.94& 60.50& 34.86& 68.00& 42.25& 80.15& 37.54& 78.04 & 32.50 & 78.14& 33.63\\
  Baseline + Bags& 89.75& 70.37& 81.56& 76.35& 72.86& 66.50& 77.93& 72.00& 79.13 & 52.21 & 73.99 & 45.61 & 74.23& 47.09\\
  Baseline + DPMA& \textbf{90.50}& \textbf{75.74}& \textbf{82.93}& \textbf{81.64}& \textbf{73.86}& \textbf{71.90}& \textbf{79.11}& \textbf{77.34}& \textbf{81.02} & \textbf{62.12} & \textbf{78.18} & \textbf{53.44} & \textbf{78.31}& \textbf{55.39}\\
  \Xhline{1.0pt}
  \end{tabular}
  \label{ablation:dmpp_sw_fmpm}
  \vspace{-2ex}
\end{table*}

\begin{table*}[t]
  \caption{
  The influence of the DPMA on retrieving continuous partial patches (CPP) and non-continuous partial patches (NCPP). (Metric: mAP)}
  \vspace{-2ex}
  \centering
  \begin{tabular}{
    m{.24\columnwidth}|
    m{.07\columnwidth}<{\centering} m{.07\columnwidth}<{\centering}|
    m{.07\columnwidth}<{\centering} m{.07\columnwidth}<{\centering}|
    m{.07\columnwidth}<{\centering} m{.07\columnwidth}<{\centering}|
    m{.07\columnwidth}<{\centering} m{.07\columnwidth}<{\centering}|
    m{.07\columnwidth}<{\centering} m{.07\columnwidth}<{\centering}|
    m{.07\columnwidth}<{\centering} m{.07\columnwidth}<{\centering}|
    m{.07\columnwidth}<{\centering} m{.07\columnwidth}<{\centering}}
  \Xhline{1.0pt}
& \multicolumn{2}{c|}{{CSVTRv2}} & \multicolumn{2}{c|}{ReCTS}& \multicolumn{2}{c|}{LSVT}& \multicolumn{2}{c|}{Avg. (Chinese)}& \multicolumn{2}{c|}{STR}& \multicolumn{2}{c|}{ArT}& \multicolumn{2}{c}{Avg. (English)} \\
  \cline{2-15}
  & CPP& NCPP& CPP& NCPP& CPP& NCPP& CPP& NCPP& CPP& NCPP& CPP& NCPP& CPP& NCPP\\
  \hline  
  Baseline & 37.64& 58.72& 46.38& 49.85& 35.31& 33.02& 41.48& 45.37& 26.38 & 50.73 & 29.14 & 37.23 & 28.56& 40.49\\
  Baseline + Bags& \textbf{75.25}& 58.59& \textbf{81.85}& 47.38& 72.59& 41.61& 77.39& 47.53& \textbf{60.99} & 41.83 & 53.23 & 34.90 & 54.87& 36.58\\
  Baseline + DPMA& 74.28& \textbf{78.24}& 81.65& \textbf{81.64}& \textbf{73.04}& \textbf{67.23}& \textbf{77.79}& \textbf{75.53}& 60.01 & \textbf{64.62} & \textbf{54.26} & \textbf{52.29} & \textbf{55.48}& \textbf{55.27}\\
  \Xhline{1.0pt}
  \end{tabular}
  \label{ablation:dmpp_sw_pmjm}
  \vspace{-2ex}
\end{table*}

\subsection{Ablation Study}
We conduct ablation studies on three Chinese datasets (~\ie, {CSVTRv2}, ReCTS, LSVT)  and two English datasets (~\ie, STR, ArT) to show the effectiveness of the proposed components including the cross-model similarity learning (CMSL), the RankMIL and the DPMA.

\subsubsection{Ablation Study on the CMSL}

We verify the effectiveness of CMSL by comparing it with PHOC-based methods{~\cite{GomezECCV20118YOLO+PHOC}} and end-to-end (E2E) recognition methods on the TIR task. For fair comparisons, we build the model of PHOC-based and E2E-based methods by adding a module for predicting PHOCs and a CTC-based recognizer{~\cite{liu2020abcnet}} for translating after the text-line proposal module.

In Tab.~\ref{abla:cmsl}, we can see that the method of ``Baseline + CMSL” achieves the best performances among all Chinese/English datasets. Compared to the method of ``Baseline + E2E”, our method outperforms it over 4.0\% mAP in average on English datasets. As the number of Chinese character type is large, the dimension of PHOC dramatically increases, making it difficult for the network to converge. These results reveal that the proposed CMSL is more robust and can be easily generalized to non-Latin scripts.

\vspace{-2ex}
\subsubsection{The Impact of the Hyperparameter $K$}

Boundary TextSpotter~\cite{boundarytextspotter} confirms that it can achieve the best performances when $K$ is 7. We also discuss how different values of $K$ affect the overall results of our scene text retrieval method. We conduct experiments on the ArT dataset and the LSVT dataset by setting $K$ as the values 3, 5, 7, and 9. From Tab.~\ref{abl_k}, first, we can see that boundary points are a better shape representation of multi-oriented/curved text than rectangles for text retrieval tasks. Then, we can conclude that retrieval performances increase when $k$ varies from 3 to 7. However, as $K$ reaches 7, the performances become stable despite larger $K$.
\vspace{2ex}

\subsubsection{Ablation Study on the RankMIL}
We evaluate the effectiveness of our proposed RankMIL approach for TIR and PPR tasks, comparing it to two other training strategies: the CMSL and the MIL. For the PPR task, we evaluate the tested models using the DPMA. The experimental results are presented in Tab.~\ref{ablation:seq_mil}.
In contrast to the CMSL, the MIL and the RankMIL training strategies require extra bag data $\{P^r, T^r\}$ as input for training. To ensure fair comparisons, we introduce the two variables to the CMSL. The ``CMSL-A” means that we train the model only by the CMSL. The ``CMSL-B” indicates that the model is trained by adding $T^r$ as extra query words. The ``CMSL-C” assumes that the $i$-th word in $T^r$ is the translation label of $i$-th proposal in $P^r$, then packages them as a labeled text-line instances to train a model by the CMSL.

\textbf{Bag Data}. Based on the “CMSL-A”, the “CMSL-B” that adds $T^r$ as query texts for training brings 1.49\% mAP performance gain in average among Chinese datasets and 1.36\% mAP performance increase in average among English datasets for the PPR task, while resulting in decreases of 0.69\% mAP and 1.03\% mAP in the TIR task. However, when adding $P^r$ to the training data, the performances of both tasks decreases across all datasets. The two experimental results are both attributable to training data. The inclusion of extra $T^r$ improves the robustness of the model for the PPR task, as it has seen enough partial query words. However, the extra $P^r$ with noisy labels may mislead the CMSL, significantly reducing the accuracy of similarity measurement.

\textbf{MIL and RankMIL}. For each query word in $T^r$, MIL first searches a proposal of the largest similarity from $P^r$, then assigns it to this query word as its text instance. Benefiting from the collaborative training with the TIR task optimized with the labeled data, the searched most similar text instance is accurate, which decreases the negative impacts of noisy samples in “CMSL-C”. MIL improves the average mAP of TIR and PPR tasks by 3.67\% and 3.52\% on Chinese datasets, and 3.53\% and 2.74\% on English datasets, respectively, compared to ``CMSL-C". Furthermore, MIL effectively utilizes bag data and outperforms ``CMSL-A" by over 2.00\% mAP on both tasks across all datasets. The proposed RankMIL further eliminates the negative impacts brought by MIL. On average, RankMIL outperforms MIL by 3.90\% mAP on Chinese datasets and 3.35\% mAP on English datasets in the PPR task. Due to decreasing the negative impact of noisy samples for similarity learning, the TIR performances are increased by more than 1.5\% mAP on Chinese datasets and English datasets.

\begin{figure}[t]
    \includegraphics[width=1.0\linewidth]{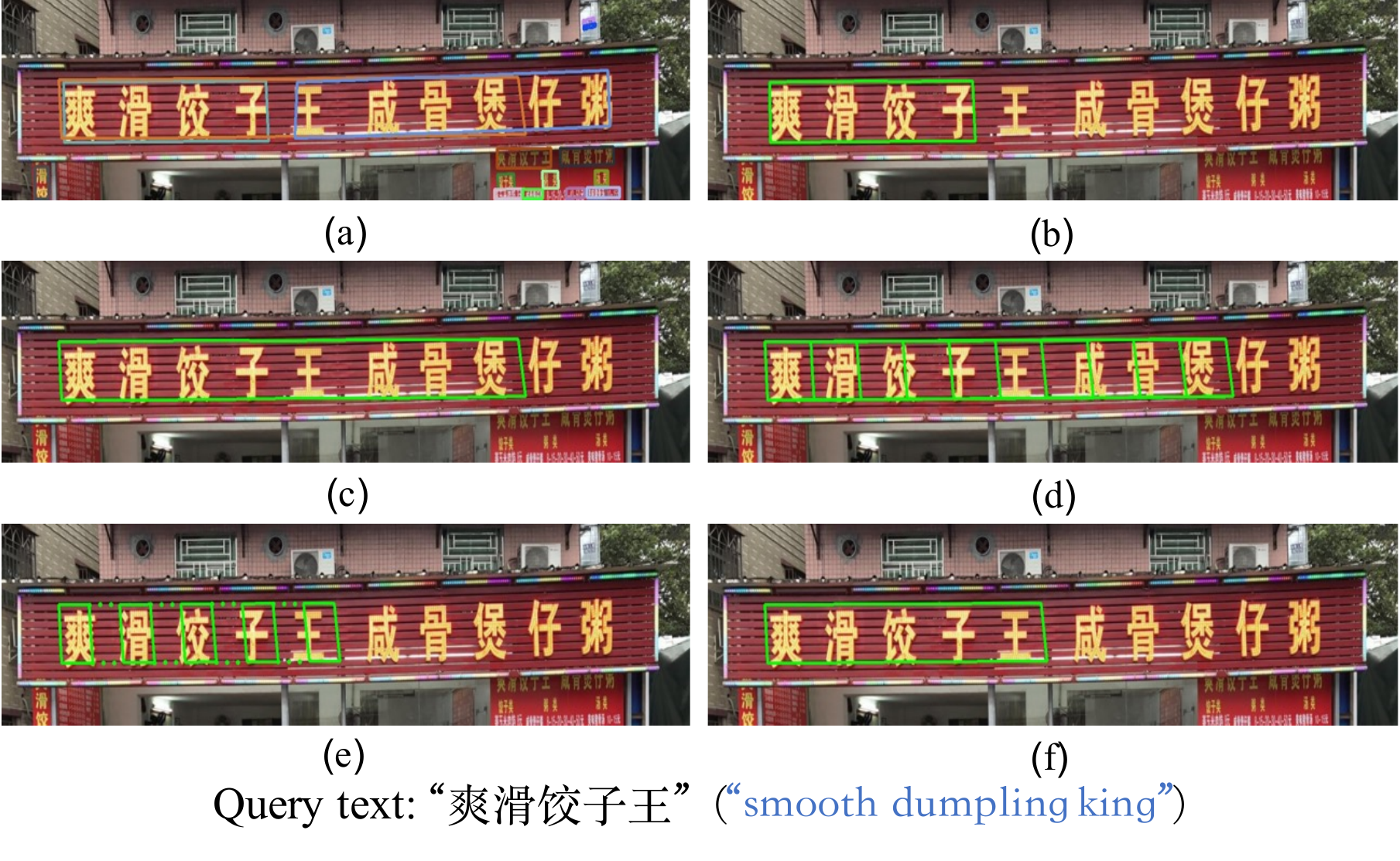}
\centering
\vspace{-5ex} 
  \caption{The reason why ``Baseline + Bags" and ``Baseline + DPMA" can promote the TIR performance of ``Baseline" in Chinese datasets. (a): detected text-line proposals; (b): the retrieved text-line proposal without using the DPMA; (c): a text-line proposal containing query text; (d): partial patches in the constructed bag; (e): the searched optimal partial patch by the DPMA in the text-line proposal (c); (f): the matched partial patch among all partial proposals in (d) by constructing bags. The word in blue is the English translation corresponding to the Chinese word in black.}
\vspace{-2ex}  
  \label{fig:dmpp_promote_fm}
\end{figure}

\begin{figure*}[t]
    \includegraphics[width=0.96\linewidth]{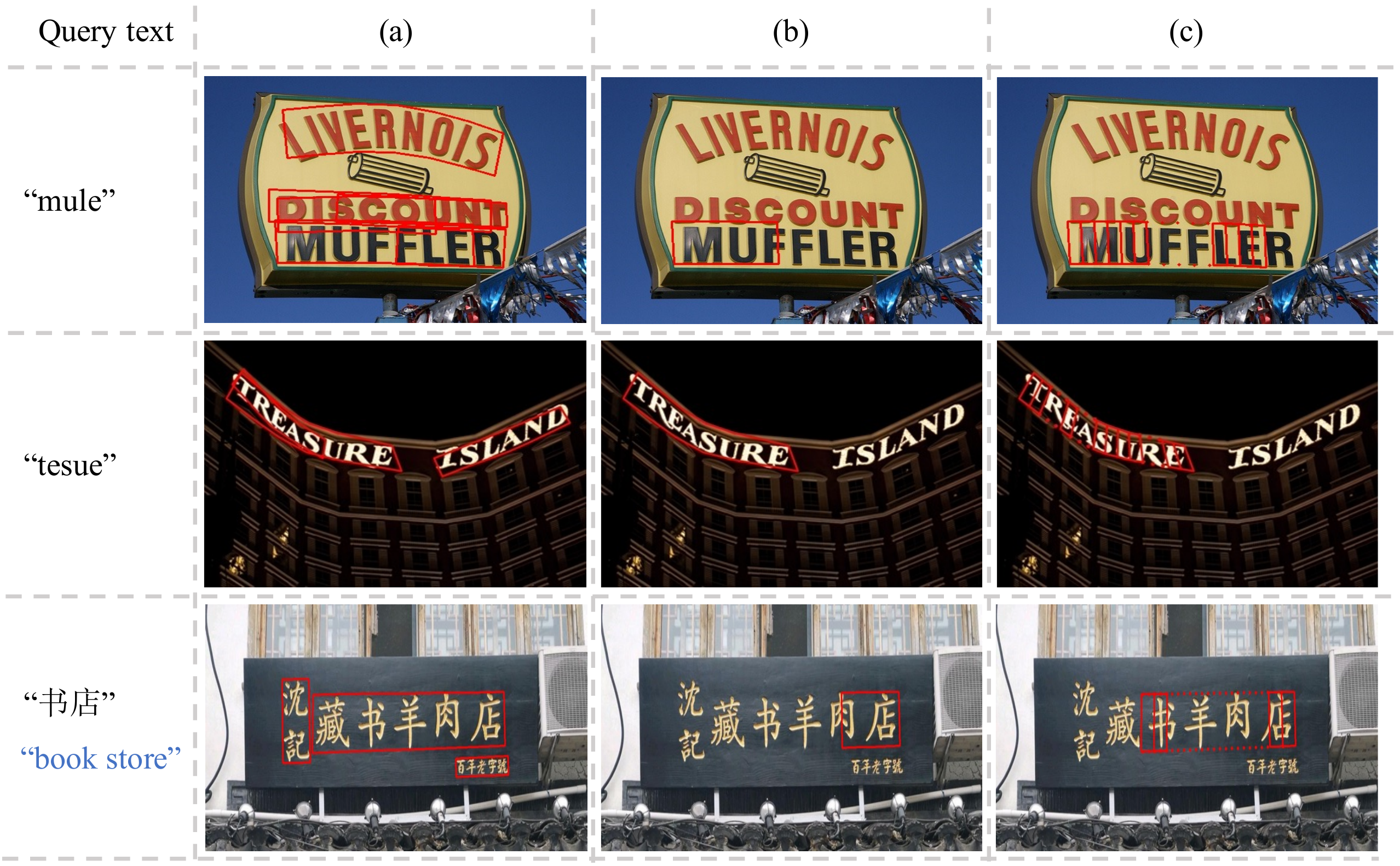}
\centering
\vspace{-2ex} 
  \caption{Comparison of retrieving non-continuous partial patches by the DPMA and constructing bags. (a): detected text-line proposals; (b): retrieved partial patches with constructing bags; (c): the searched optimal partial patches by the DPMA. The segments linked by the dashed lines compose the non-continuous partial patches that does not contains middle parts covered by the dash lines. The word in blue is the English translation corresponding to the Chinese word in black.}
  \label{fig:dmpp_sw}
\end{figure*}

\textbf{The Hyper-parameter $m$ in the RankMIL}.
The margin $m$ is the only hyper-parameter of RankMIL. We show its effect on the retrieval performances in Fig.~\ref{fig:margin}. All experiments are conducted on the ReCTS dataset. As $m$ approaches 0, the performances of TIR and PPR are close to those of MIL (80.78\% and 77.45\% mAP), indicating that the RankMIL without filtering noisy samples degenerates into the MIL. Moreover, the performances improve as $m$ increases and reach maximum values when $m$ equals 0.2. Those experimental results confirm the effectiveness of RankMIL. In other experiments, we set $m$ to 0.2.

The experimental results reveal four conclusions. Firstly, bag data is essential for training a robust model for the PPR task. Secondly, when labeled partial patches are missing in the training dataset, a conventional MIL outperforms the CMSL as a training strategy. Thirdly, the PPR performance can be further enhanced by eliminating noisy samples in MIL. Fourthly, an effective training strategy for the PPR task can improve the TIR performance, otherwise hurt its retrieval accuracy.

\subsubsection{Ablation Study on the DPMA}

As illustrated in Alg.~\ref{alg:inference}, generating bags by BCA is a feasible solution to the PPR. We compare our proposed DPMA with the method of generating bags. We first evaluate the two methods for TIR and PPR tasks. Then, we compare their performances of retrieving continuous/non-continuous partial patches. All evaluated models are trained by the RankMIL.

\textbf{On TIR and PPR Tasks}.  
The experimental results presented in Tab.~\ref{ablation:dmpp_sw_fmpm} show that the baseline method achieves competitive performance for the TIR task across all datasets without the use of bags or the DPMA. However, the PPR performances are significantly lower. The introduction of bags or the DPMA leads to an improvement in the PPR performance of approximately 30\% mAP. Furthermore, the use of the DPMA also promotes the TIR task across all datasets, particularly on the more challenging Chinese datasets (~\ie, ReCTS and LSVT), where text detection is more difficult. In cases where the detected text-line proposals are not accurate enough, the DPMA and the method with bags can search for a tighter proposal that covers the text-line instance. As the shape of scene text instances in the ArT dataset is complex, the constructed bag contains many noisy patches. This results in a decrease in the TIR performance. In addition, for the PPR task, the DPMA outperforms the method with bags in average by 5.34\% mAP on Chinese datasets and 8.30\% mAP on English datasets.

As shown in Fig.~\ref{fig:dmpp_promote_fm}, no detected text-line proposal in (a) can cover a text instance translated as the given query, resulting in an incorrect retrieval result in (b). However, in (d), the approach with bags can produce numerous partial patches from the text-line proposal in (c), followed by the selection of the optimal partial patch in (f). In contrast, the DPMA can directly search for the optimal partial patch in (e) from text-line proposal (c) without constructing a bag. Furthermore, compared to (f), the optimal partial patch in (e) contains less background noise, thereby enabling more precise similarity measurement between the query text and the partial patch.

\begin{table*}[t]
  \caption{
  Performance comparisons with the state-of-the-art text retrieval/spotting methods on STR, CTR and ArT. (Metric: mAP)}
  \vspace{-2ex}
  \centering
  \begin{tabular}{
    m{.35\columnwidth}|
    m{.1\columnwidth}<{\centering} m{.1\columnwidth}<{\centering}|
    m{.1\columnwidth}<{\centering} m{.1\columnwidth}<{\centering}|
    m{.1\columnwidth}<{\centering} m{.1\columnwidth}<{\centering}|
    m{.1\columnwidth}<{\centering} m{.1\columnwidth}<{\centering}}
  \Xhline{1.0pt}
  \multirow{2}*{Method}& \multicolumn{2}{c|}{STR} & \multicolumn{2}{c|}{CTR}& \multicolumn{2}{c|}{ArT}& \multicolumn{2}{c}{Avg.}  \\
  \cline{2-9}
  & TIR& PPR& TIR& PPR& TIR& PPR& TIR& PPR\\
  \hline
  Mishra~\etal~\cite{MishraAJ13} & 42.70& -&- & -&-& -&-& -\\
  Jaderberg~\etal~\cite{jaderberg2016reading}& 66.50& -&- & -&-& -&-& -\\
  He~\etal~\cite{HeCVPR2018}&  66.95& -&- & -&-& -&-& -\\
  Gomez~\etal~\cite{GomezECCV20118YOLO+PHOC}& 69.83& -& 41.05& -& 55.28& -& -& -\\
  ABCNet~\cite{liu2020abcnet}& 67.25& 33.09& 48.33& 29.48& 64.56 & 30.56& 59.48& 30.61\\
  Mask TextSpotter v3~\cite{liao2020mask} & 74.48& 45.59& 55.54& 35.84& 70.72& 40.32& 66.01& 39.72\\
  SwinTextSpotter~\cite{swintextspotter} & 76.66& 48.46& 53.19& 40.75& 75.51& 41.43& 68.45& 42.30\\
  
  Conference version~\cite{Wang_2021_CVPR}& 77.09& 33.13& 66.45& 30.32& 70.00& 30.56& 69.10& 30.88\\
  Ours  & 80.15& 37.54& \textbf{72.95}& 36.41& 77.94& 32.50& 76.42& 34.50\\
  Ours (with DPMA)& \textbf{81.02}& \textbf{62.12}& 72.57& \textbf{54.18}& 
 \textbf{78.18}& \textbf{53.44}& \textbf{76.49}& \textbf{55.01} \\
  \Xhline{1.0pt}
  \end{tabular}
  \label{sota:english}
  \vspace{+2ex}
\end{table*}

\begin{figure*}[t]
    \includegraphics[width=1.0\linewidth]{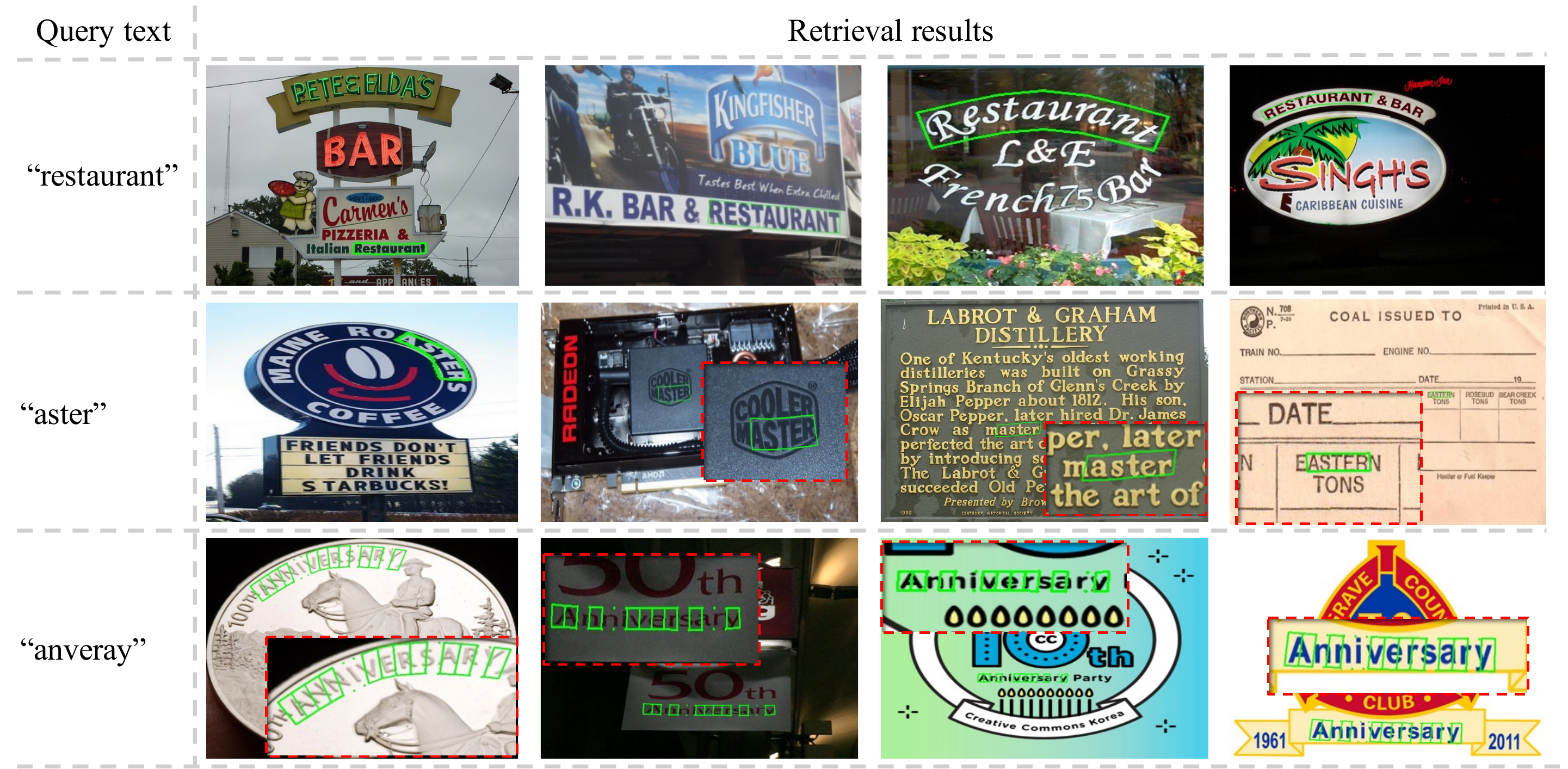}
\centering
\vspace{-4ex}
  \caption{Visualization of the retrieval results. The three-row examples are from the results of TIR, PPR (CPP), and PPR (NCPP), respectively.}
  \label{fig:result_sr_en}
\end{figure*}

\textbf{On Retrieving CPP and NCPP}. 
In Tab.~\ref{ablation:dmpp_sw_pmjm}, the approach with bags achieves substantial improvements for the CPP but limited performances for the NCPP, as bags only contain continuous partial patches. In contrast, the proposed DPMA delivers an average mAP improvement of approximately 30\% for both CPP and NCPP across three Chinese datasets, narrowing the performance gap between handling the two types of partial patches. These experimental results highlight the superiority of DPMA. As shown in Fig.~\ref{fig:dmpp_sw}, when the query text is non-continuous one of the text-line proposal in (a), the DPMA successfully localizes and retrieves NCPP (as in (c)), while the method with bags only probes continuous partial patches (as in (b)) from the text-line instances. The competence of DPMA to NCPP makes our approach applicable to a broader range of scenarios.

\begin{table*}[t]
  \caption{
  Performance comparisons with the state-of-the-art text retrieval/spotting methods on {CSVTRv2}, ReCTS and LSVT datasets. (Metric: mAP)}
  \vspace{-2ex}
  \centering
  \begin{tabular}{
    m{.35\columnwidth}|
    m{.1\columnwidth}<{\centering} m{.1\columnwidth}<{\centering}|
    m{.1\columnwidth}<{\centering} m{.1\columnwidth}<{\centering}|
    m{.1\columnwidth}<{\centering} m{.1\columnwidth}<{\centering}|
    m{.1\columnwidth}<{\centering} m{.1\columnwidth}<{\centering}}
  \Xhline{1.0pt}
  \multirow{2}*{Method}& \multicolumn{2}{c|}{{CSVTRv2}} & \multicolumn{2}{c|}{ReCTS}& \multicolumn{2}{c|}{LSVT}& \multicolumn{2}{c}{Avg.}  \\
  \cline{2-9}
  & TIR& PPR& TIR& PPR& TIR& PPR& TIR& PPR\\
  \hline  
  ABCNetv2~\cite{journals/corr/abs-2105-03620}& 78.93& 44.95& 68.33& 49.23& 37.53& 23.92& 54.66& 39.28\\
  Conference version~\cite{Wang_2021_CVPR}& 86.20& 42.86& 70.31& 45.92& 58.31& 32.76& 65.51&	40.65\\
  Ours (E2E)& 86.06& 49.50& 70.48& 51.33& 55.26& 33.18& 64.19& 44.33\\
  Ours   & 89.45& 45.42& 73.00& 46.94& 60.50& 34.86& 68.00& 42.25\\
  Ours (with DPMA)&\textbf{90.50}& \textbf{75.74}& \textbf{82.93}& \textbf{81.64}& \textbf{73.86}& \textbf{71.90}&  \textbf{79.11}& \textbf{77.34}\\
  \Xhline{1.0pt}
  \end{tabular}
  \label{sota:chinese}
  \vspace{+3ex}
\end{table*}

\begin{figure*}[t]
    \includegraphics[width=0.99\linewidth]{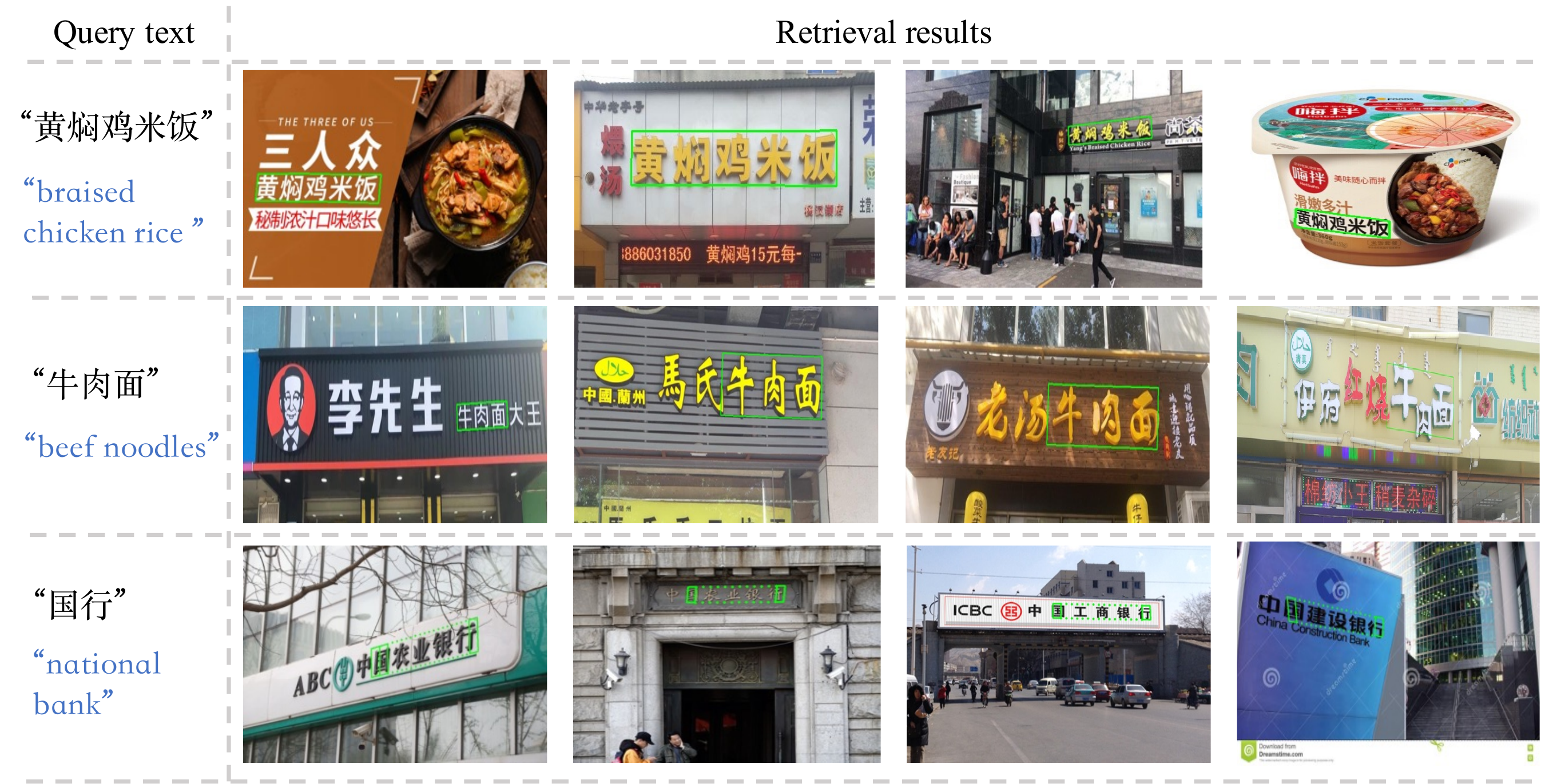}
\centering
\vspace{-1ex} 
  \caption{Visualization of the retrieval results. The three-row examples are from the results of TIR, PPR (CPP), and PPR (NCPP), respectively. The words in blue are the English translations corresponding to the Chinese words in black.}
\vspace{-2ex}  
  \label{fig:result_sr_cn}
\end{figure*}

\vspace{-1.5ex}
\subsection{Comparisons with Previous Methods}

We compare our proposed method with previous retrieval/spotting methods on three English datasets and three Chinese datasets. For the spotting methods, edit distances between a query word and the spotted words from scene images are used for image retrieval.

\subsubsection{Experimental Results on English Datasets}
The experimental results of previous state-of-the-art methods and our method are summarized in Tab.~\ref{sota:english}.
We also compare our method to the three most recent state-of-the-art scene text spotting approaches: Mask TextSpotter v3~\cite{liao2020mask}, ABCNet~\cite{liu2020abcnet} and SwinTextSpotter~\cite{swintextspotter} are also adopted in the comparisons. The retrieval results by Mask TextSpotter v3, ABCNet and SwinTextSpotter are obtained with their officially released models\footnote{https://github.com/MhLiao/MaskTextSpotterV3}$^,$\footnote{https://github.com/Yuliang-Liu/bezier\_curve\_text\_spotting}$^,$\footnote{https://github.com/mxin262/SwinTextSpotter}.

\textbf{On the TIR Task}. We can observe that ``Ours (with DPMA)" outperforms other methods by significant margins over all datasets.  
While SwinTextSpotter~\cite{swintextspotter} achieves the best average performance among previous approaches due to its stronger backbone Swin-Transformer~\cite{swintransformer}, it still falls short of our conference version~\cite{Wang_2021_CVPR} by 0.65\% mAP. However, our conference version performs much worse than SwinTextSpotter on the ArT dataset because its text detection module can only handle horizontal text instances. By adopting the representation of boundary points~\cite{boundary}, our method outperforms SwinTextSpotter by 2.43\% mAP on the ArT dataset. Compared to the previous method with the same Resnet-50 backbone, our method achieves an average performance gain of more than 10.0\% mAP over Mask TextSpotter v3~\cite{liao2020mask}. 
The consistent improvements over both PHOC-based methods and end-to-end text spotters demonstrate the importance of learning cross-modal similarity for scene text retrieval.

\textbf{On the PPR Task}. 
Except for ``Ours (with DPMA)", SwinTextSpotter~\cite{swintextspotter} achieves the best average performance, as a sub-sequence of a word can be found by string matching from the recognized text. Yet, for SwinTextSpotter, the performance gap between the TIR and PPR tasks is still over 25.0\% mAP. The gap indicates that the PPR task has a higher performance requirement for the spotting methods, as a wrongly recognized character more easily changes the sub-sequence of a word. ``Ours (with DPMA)" outperforms SwinTextSpotter by 12.71\% mAP and reduces the gap to around 20.0\% mAP on average among the three datasets.

Some qualitative results of our method are shown in Fig.~\ref{fig:result_sr_en}. We can see that the proposed method can accurately localize and retrieve scene text instances including text-line instances, continuous partial patches, and non-continuous partial patches.
In the last image of this figure, we can see that the character 'A' occupies two units while the smaller character 'y' occupies only one unit. This indicates that the DPMA can reduce the negative impact of scale variations to a certain extent.

\subsubsection{Experimental Results on Chinese Datasets}
The most recent state-of-the-art Chinese scene text spotting approaches, namely ABCNet v2~\cite{liao2020mask} is adopted in the comparisons. Besides, we also build a CTC-based recognition head on the top of our text-line proposal module to form a text spotter for comparisons. The retrieval result by ABCNet v2 is obtained with their officially released models\footnote{https://github.com/aim-uofa/AdelaiDet/tree/master/configs/BAText}.

From Tab.~\ref{sota:chinese}, we can see that our proposed method achieves state-of-the-art performances on all datasets. Despite ABCNet v2 can deal with arbitrary-shaped scene texts, our method of conference version~\cite{Wang_2021_CVPR} that only formulates the shape of scene text as horizontal boxes outperforms ABCNet v2 more than 10\% mAP on the TIR task. Additionally, under the same detection module and training data, ``Ours" still outperforms ``Ours (E2E)" by 3.81\% mAP. These consistent experimental results confirm that learning the cross-modal similarity is a better strategy than spotting methods for retrieving non-Latin text. 

Equipped with the DPMA, our method enhances the performance of TIR and PPR tasks by over 10\% mAP and 30\% mAP, respectively. In English scene text, different characters often occupy distinct sizes, with ``W" being much wider than ``i" in a scene image. Consequently, the performance gap of ``Ours (with DPMA)" persists between the two tasks on English datasets, as the model struggles to extract features of small-sized characters. Conversely, Chinese characters are roughly of the same size. Thus, ``Ours (with DPMA)" considerably narrows the performance gap between the two tasks, while the spotting methods continue to face a significant gap. Those experimental results further confirm the superiority of feature matching to string matching in terms of retrieval accuracy, when measuring the cross-modal similarity between a scene text instance and a word.
Some qualitative results of our method are shown in Fig.~\ref{fig:result_sr_cn}.

\vspace{-2ex}

\begin{table}[t]
  \caption{
  The influence of the DPMA in terms of inference speed and accuracy. {SwinTextSpotter is evaluated with the public model. SwinTextSpotter* is trained with our re-divided training set of ReCTS.}}
  \vspace{-2ex}
  \centering
  \begin{tabular}{
    m{.32\columnwidth}|
    m{.1\columnwidth}<{\centering} m{.1\columnwidth}<{\centering}|
    m{.1\columnwidth}<{\centering} m{.1\columnwidth}<{\centering}
    }
  \Xhline{1.0pt}
  \multirow{2}*{Method}& \multicolumn{2}{c|}{Speed (FPS)}& \multicolumn{2}{c}{Accuracy (mAP)}\\
  \cline{2-5}
  & Online& Offline& TIR& PPR\\
  \hline
  SwinTextSpotter~\cite{swintextspotter}& 5.1& 713.4& 78.79& 58.44\\
  SwinTextSpotter*~\cite{swintextspotter}& 5.4& 726.4& 71.43& 52.17\\
  ABCNetv2~\cite{journals/corr/abs-2105-03620}& 9.2& \textbf{832.1}& 68.33& 49.23\\
  Conference version~\cite{Wang_2021_CVPR}& \textbf{10.1}& 330.2& 70.31& 45.92\\
  Ours (with Bags)& 2.1& 70.2& 81.56& 76.35\\
  Ours (with DPMA)& 9.3& 313.6& \textbf{82.93}& \textbf{81.64}\\
  \Xhline{1.0pt}
  \end{tabular}
  \label{inference_speed}
  \vspace{-2ex}
\end{table}

\subsection{Inference Speed Analysis}
We compare the inference speed of our approach with previous spotting/retrieval methods on ReCTS dataset.  
The inference speed is tested in terms of online and offline manner. The online speed includes the time consumption of backbone feature extraction, network predictions (text recognition of spotting methods or text feature extraction of our method), and similarity measurement (text matching of spotting methods or cosine similarity of our method). The offline speed only includes the time of similarity measurement, assuming network predictions has been obtained beforehand. We measure the inference speed using the FPS (frames per second) metric for handling a single query word. 

{We compare our method with SwinTextSpotter in two experimental setups. SwinTextSpotter is evaluated using the public model trained on the original ReCTS training set. However, since the original ReCTS test set lacks annotations, our ReCTS test images are sampled from its training set. To ensure a fair comparison, we re-trained the SwinTextSpotter* model on our re-divided ReCTS training set using the officially released code.}
ABCNetv2~\cite{journals/corr/abs-2105-03620} and our approach have the same text detection architecture, differing only in their text shape prediction heads.

Tab.~\ref{inference_speed} shows that ABCNetv2 and ``Ours (with DPMA)" achieve comparable online speeds. However, ``Ours (with Bags)" incurs a significant drop in online speed as it involves extracting text features of more instances. In terms of offline inference speed, ABCNetv2 outperforms retrieval methods due to only requiring word matching, but it yields lower accuracy. Among the retrieval methods, ``Ours (with DPMA)" achieves the best accuracy in TIR and PPR tasks. However, ``Ours (with Bags)" suffers from much lower offline speed due to feature matching among a larger number of instances. Our approach with DPMA achieves the optimal speed-accuracy trade-off among these methods. 

The retrieval speed can decrease when text instances are extremely dense. For the scene text dataset, the inference pipeline takes approximately 100 ms per image.
In contrast, for document images, such as PDF pages, the process takes about 350 ms.

\vspace{-1ex}
\section{Limitation}
One limitation of our method is that it is difficult to deal with query text of rarely-used characters, especially for Chinese text. Although the synthetic images containing those characters are helpful to train the model, it fails when the real training data rarely contains those characters. This is a common limitation for scene text retrieval methods. {
Besides, when query text is extremely long, the retrieval performance significantly drops.
}
\vspace{-1ex}
\section{Conclusion}
In this paper, we have presented a partial scene text retrieval framework that combines scene text detection and similarity learning to search for text instances that match or are similar to a given query text from images. Moreover, the proposed approach can retrieve text-line instances and partial patches without extra annotations, which is a more challenging task for existing approaches. Our experiments demonstrate that our method consistently outperforms state-of-the-art retrieval/spotting methods on English and Chinese datasets. In the future, we aim to extend this method to handle more complex cases, such as synonyms.

\vspace{-1ex}
\section*{Acknowledgements}
This work was supported by the National Science Fund for Distinguished Young Scholars of China Grant No.62225603 and the National Science and Technology Major Project under Grant No.2023YFF0905400.

\ifCLASSOPTIONcaptionsoff
  \newpage
\fi

\bibliographystyle{IEEEtran}
\bibliography{reference}

\end{document}